\documentclass{article} 
\usepackage{iclr2027_conference,times}

\usepackage{amsmath,amsfonts,bm}

\def\eqref#1{equation~\ref{#1}}

\def\1{\bm{1}}

\DeclareMathAlphabet{\mathsfit}{\encodingdefault}{\sfdefault}{m}{sl}
\SetMathAlphabet{\mathsfit}{bold}{\encodingdefault}{\sfdefault}{bx}{n}

\usepackage{hyperref}
\usepackage{url}

\title{\method: Agentic Seam Planning for \\UV Parameterization}

\author{Marco Schouten, Arthur Roullier, Élie Michel, Ruben Wiersma, Axel Paris, Tamy Boubekeur\\
Adobe Research, France\\
\texttt{\{mschouten,roullier,emichel,rwiersma,aparis,boubek\}@adobe.com} \\
}

\iclrfinalcopy 

\newcommand{\method}{\textsc{MeshQuery}}
\newcommand{\story}[1]{\noindent\textcolor{gray}{\textit{Story:} #1}\par}
\newcommand{\coremessage}[1]{\noindent\textbf{Core message:} #1\par\medskip}

\usepackage{amsmath}
\usepackage{amssymb}
\usepackage{mathtools}
\usepackage{bm}

\usepackage{graphicx}

\usepackage{booktabs}
\usepackage{multirow}
\usepackage{tabularx}
\usepackage{array}
\usepackage{siunitx}

\usepackage{microtype}
\usepackage{xspace}

\usepackage{enumitem}

\usepackage{xcolor}

\usepackage{algorithm}
\usepackage{algorithmic}

\usepackage{xcolor}
\usepackage[normalem]{ulem}

\usepackage{booktabs}
\usepackage{graphicx}
\usepackage{xcolor}
\usepackage{colortbl}

\definecolor{BenchmarkBlue}{RGB}{55,90,125}
\definecolor{BenchmarkBlueLight}{RGB}{220,232,244}
\definecolor{BenchmarkGrayLight}{RGB}{232,232,232}

\usepackage{comment}
\usepackage{cleveref}

\newcommand{\mypar}[1]{\vspace{0mm}\noindent\textbf{#1}}
\newif\ifshowcomments
\showcommentstrue  

\definecolor{LlmColor}{RGB}{100,100,100}
\newcommand{\llm}[1]{\textcolor{LlmColor}{#1}} 

\definecolor{ElieColor}{RGB}{170,46,200}
\newcommand{\elie}[1]{\ifshowcomments\textcolor{ElieColor}{\textit{(É: #1)}}\fi} 
\newcommand{\elieUpdate}[1]{\textcolor{ElieColor}{#1}} 

\definecolor{TamyColor}{RGB}{148,0,148}

\definecolor{RubenColor}{RGB}{46,200,170}

\definecolor{ArthurColor}{RGB}{7,27,247}
\newcommand{\arthur}[1]{\ifshowcomments\textcolor{ArthurColor}{(A: #1)}\fi}
\newcommand{\arthurUpdate}[1]{\textcolor{ArthurColor}{#1}}

\definecolor{MarcoColor}{RGB}{230,120,20}
\newcommand{\marco}[1]{\ifshowcomments\textcolor{MarcoColor}{M: #1}\fi}
\newcommand{\marcoUpdate}[1]{\textcolor{MarcoColor}{#1}}

\definecolor{AxelColor}{RGB}{90,170,15}

\newcommand{\axelUpdate}[1]{\textcolor{AxelColor}{#1}}

\definecolor{WarnColor}{RGB}{220,25,15}

\newcommand\inlinecode[1]{\texttt{\small #1}}

\newcommand\evaluator{I}
\newcommand\agent{A}
\newcommand\seamprog{P}
\newcommand\seamplan{S}
\newcommand\uvatlas{U}
\newcommand\metrics{q}
\newcommand\edges{E}
\newcommand\VLM{VLM} 
\newcommand\aVLM{a VLM} 
\newcommand\VLMfull{Vision-Language Model} 
\newcommand\inputmesh{M}
\newcommand\meshdb{M^*}
\newcommand\processor{\Phi}
\newcommand\iteration{n}

  \usepackage{amsmath}
  \usepackage{booktabs}
  \usepackage{longtable}
  \usepackage{array}

\begin{document}

\maketitle
\lhead{}

\begin{figure}[h!]
    \centering
    \includegraphics[width=\linewidth]{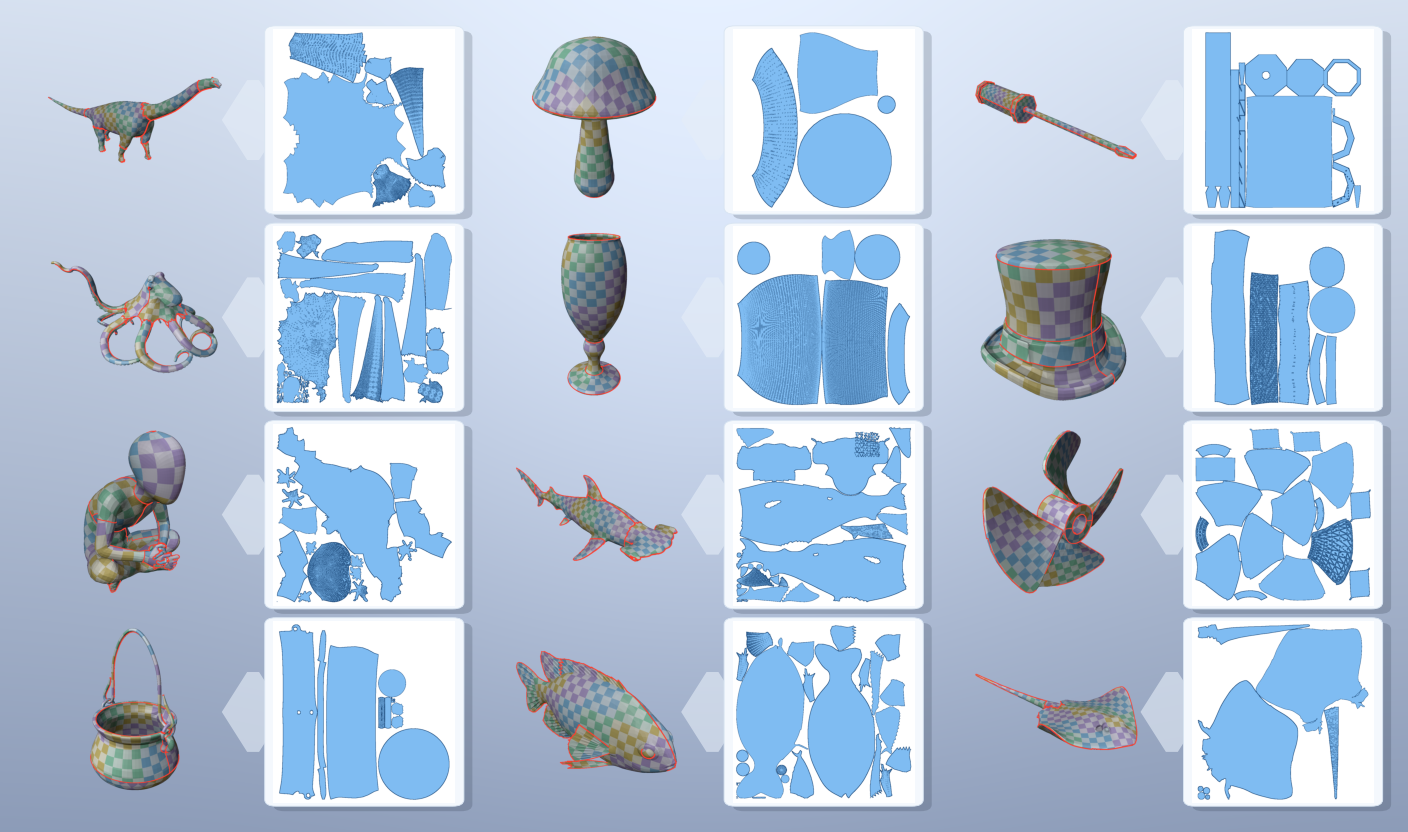}
\caption{\textbf{\method{} generates artist-aligned UV seams.}
Conversely to hand-crafted geometric optimization, \method{} enables a \VLM{} agent
to plan seams the way an artist does: it iteratively refines seams
expressed as programs in a Domain-Specific Language (DSL), based on unwrapping feedback and on the mesh's geometric, topological, semantic attributes, and downstream needs.}
  \label{fig:teaser}
\end{figure}

\begin{abstract}
We present \method{}, a training-free agentic approach to automatic UV unwrapping of production-grade quad meshes. A \VLMfull{} (\VLM{}) plans artist-aligned seams using a set of edge-selection tools, conditioned on domain-specific UV-unwrapping knowledge expressed in natural language and refined with a feedback loop.
We design a queryable mesh representation together with a domain-specific language (DSL) that enables the agent to retrieve mesh information on demand, express a seam plan as a compact program of edge-selection operators over topological, geometric, and semantic mesh attributes, and iteratively refine it from UV quality feedback.
On Adobe Substance 3D and Toys4K meshes, \method{} produces $2.9\times / 4.29\times$ fewer charts and $1.63\times/1.7\times$ shorter seams than the strongest baseline, and professional artists prefer its results in 80.9\% of comparisons. Ultimately, decoupling high-level intent planning from low-level edge selection and compact mesh representation let \method{} run on different backend VLM and  scale to meshes an order of magnitude larger than autoregressive seam prediction.
\end{abstract}

\section{Introduction} 
\label{sec:introduction}


Surface parameterization plays a key role in 3D modeling pipelines, for example, for texture mapping, mesh editing, decaling, and fabrication~\cite{sheffer2006survey}. In most cases, the surface comes in the form of a 2-manifold polygon mesh, potentially open and/or exhibiting multiple connected components, and is first cut along seams to form components with disk topology, reduce distortion, and ensure a bijective mapping with the 2D domain, e.g., texture space. In this work, we focus on \textit{seam planning}, i.e., the selection of edges to mark as seams. Seam planning is difficult to automate because it combines high-level considerations (semantics, downstream usage) and low-level geometric considerations (distortion, number of islands), which are hard to satisfy simultaneously. Moreover, high-level considerations are hard to quantify (semantics, symmetry) and
often depend on the use case and type of shape. In practice, none of the existing automatic seam placement solutions \cite{wang2025partuv, xatlas, blender_smart_uv, zhang2024flatten, li2018optcuts} handle all cases well and recent data-driven approaches~\cite {li2025autoregressive, xu2025seamcrafter, xu2026seamgen, ma2026meshtailor} struggle to scale to larger meshes and require training on a dataset of high-quality seams, which is difficult to acquire. Therefore, UV unwrapping remains a manual, time-consuming process.
In this paper, we introduce \method, an agentic framework for planning seams on quad meshes built around a \VLMfull{} (\VLM{}) equipped with domain specific tools.
As we focus on the agentic loop, we limit our scope to quad-meshes, where multi-edge selection tools are readily available and of high quality.
A core challenge is how the \VLM{} interfaces with the input mesh. Interfacing directly with mesh vertices and edges is costly, especially for large meshes. An alternative is to encode the mesh into the \VLM{}’s latent space. We avoid this route, as it requires an additional training stage for the mesh encoder and is dependent on a specific \VLM{}. Our solution works directly with the existing modalities of the \VLM{}, namely text and images. Building on the \VLM{}'s ability to write programs, we represent the seam plan as a seam-selection program, for which we design a Domain Specific Language (DSL). The DSL expresses selection of edges by combining geometry (e.g., edge visibility, length) and connectivity (e.g., edge loops). The DSL additionally includes an active perception mechanism which selectively requests information from the mesh, allowing the method to scale to large meshes. The DSL lets us take advantage of the \VLM{}'s ability to reason about the task at a high level, while grounding the seam plan in precise, mesh-specific edge selection.




\mypar{Contributions.} We propose a novel agentic UV unwrapping framework 
able to generate high-quality 2D parameterizations on quad meshes (both organic and hard-surfaces) that can follow production-specific constraints, adapt to asset-specific requirements, and scale to meshes of more than $300k$ faces. We introduce a representation of the input mesh that is better suited for active and UV-seam centric perception by a \VLM{} agent, and a Domain Specific Language to express seam selections in a compact and scalable way. We also report how we refined the system prompt (skill file) of our agentic system to improve surface parameterization quality, as an example of how individual artists may adapt the system to their needs. We demonstrate that \method achieves better results than state-of-the-art, both qualitatively through a user study and quantitatively using common metrics such as distortion, island count, and UV occupancy.

\section{Related Work}
\label{sec:related-work}

\mypar{Optimization-driven seam placement} aims to find seams that minimize a user-specified objective, typically minimal seam length and surface distortion. Some include additional measures, such as seam visibility \cite{sheffer2002seamster}.
Early works first compute a set of seams using heuristics (e.g., cutting through high-curvature regions or clustering normals) and unwrap separately~\cite{sander2001texture, levy2002, guGeometryImages2002a, zhou2004isocharts, julius2005dcharts,lucquin2017seamcut}.
Later works optimize seam length and distortion simultaneously, which they achieve by formulating a differentiable energy function \cite{poranne2017autocuts, sharp2018variational}, representing the cutting and mapping functions with neural networks for differentiable optimization \cite{zhang2024flatten, srinivasan2024nuvo}, or using specialized optimizers~\cite{sorkine2002bounded, li2018optcuts,  pietroni_computational_2022}.
The optimized objectives typically do not account for artistic considerations such as semantics. In our approach, we expect the \VLM{} to reproduce these considerations from its training data (e.g., UV-unwrapping tutorials) and prompting (skill files) and make decisions based on these considerations through its vision- and language understanding. At a low level, our approach uses heuristic tools to mark seams.

\mypar{Data-driven methods} place seams based on examples of high-quality seams. Graph neural networks~\cite{teimury2020graphseam,xu2026seamgen} and autoregressive methods~\cite{li2025autoregressive,xu2025seamcrafter,chen2025artuv,ma2026meshtailor} struggle with large meshes and are bound to the quality of the dataset they are trained on~\cite{deitke2023objaverse,zhang2025texverse}. Because our agent operates on a high-level planning stage and uses tools for the low-level seam selection, we can scale to meshes with $>300K$ vertices.
PartUV~\cite{wang2025partuv} circumvents the need for seam-specific training data using part segmentation~\cite{liu2025partfield}. We exploit implicit knowledge about seam placement in \VLM{} rather than explicit seam-placement data. Like PartUV, we perform low-level seam placement with existing tools, but use a wider, different set of tools in an agentic feedback loop.

\mypar{3D representations for LLMs} 
encode geometry into tokens that LLMs can process. Prior work serializes meshes into language-compatible tokens~\cite{wang2024llama,fang2025meshllm} or projects 3D features to internal LLM representations~\cite{xu2024pointllm, hong20233d, deng20253d}. However, encoding high-resolution input meshes is costly, and projections to \VLM's latents are tied to the corresponding \VLM{}. Instead, we query task-relevant mesh information \textit{on demand} for efficient reasoning and refinement. 
Additionally, our framework supports any backend \VLM{}.
This principle is grounded in retrieval-augmented generation (RAG)~\cite{lewis2020retrieval, asai2024self}, which selectively retrieves task-relevant information rather than placing all available evidence in context.


\mypar{Agentic interface design} determines how agents perceive and interact with an environment. \cite{yang2024swe} and \cite{wang2024executable} show that the representation of \textit{observations} and \textit{actions} can substantially affect performance. Low-level interactions are usually abstracted through structured skills, programs, tools, and planners, enabling reasoning over higher-level task~\cite{ahn2022can, huang2022inner, singh2022progprompt, liu2023llm, yao2022react, schick2023toolformer, shinn2023reflexion, wang2023voyager, wu2023autogen}. Agents for computer use often involve specific point-and-click interfaces \cite{zhang2026psbench,agashe2025agent}. However, placing seams in 3D requires complex camera placement to enable point-and-click selection. We circumvent this problem by designing an efficient relational database interface for queries over mesh properties.

\section{Method }
\label{sec:method}

\begin{figure*}[t]
  \centering
  \includegraphics[width=\linewidth]{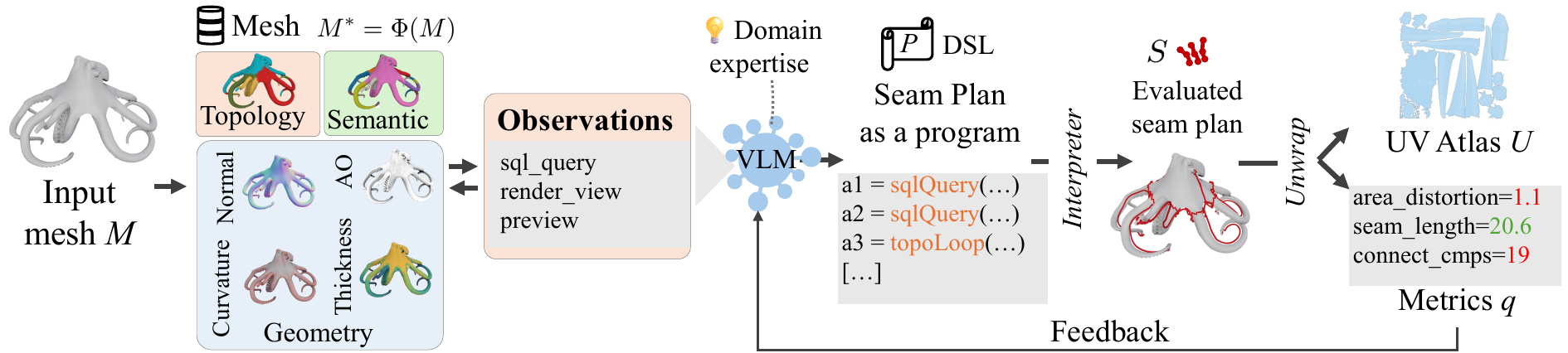}
  \caption{\textbf{Agentic UV seam planning.} The input mesh $M$ is first
converted into a queryable representation $\meshdb$ that stores geometric,
topological, and semantic attributes. Then a \VLM{} agent, conditioned on
seam-preference knowledge, iterates a three-stage loop: it \emph{observes}
$\meshdb$ through SQL queries and seam-selection statement previews, then generates a cohesive program $\seamprog$, evaluates the plan through its UV quality metrics $\metrics$, and iteratively refines it until all quality bounds are met.}
  \label{fig:method-pipeline}
\end{figure*}


We propose an agentic framework to mark seams on a mesh for UV unwrapping.
Our framework (illustrated in Fig.~\ref{fig:method-pipeline}) is an iterative loop where \aVLM{} invokes fixed functions called \textit{tools} and processes their responses. 
First, we build an efficient representation of the input mesh $\meshdb = \Phi(\inputmesh)$ that \aVLM{} can perceive using observation tools that query mesh semantics, geometry, and topology (Sec.~\ref{sec:mesh_db}). Second, we define a Domain-Specific Language (DSL) that enables \aVLM{} to mark seams via aggregate seam-selection statements that operate on $\meshdb$ (Sec.~\ref{sec:selection_dsl}). The \VLM{} previews each statement on the geometry through a DSL interpreter and composes a joint \textit{seam selection program} $\seamprog$, which the interpreter instantiates as a seam plan $\seamplan$. Third, a fixed parameterization and packing operator $\Psi$ processes $\seamplan$ to produce the UV atlas, and we evaluate quality metrics $q$ that we return to the agent as feedback (Sec.~\ref{sec:submit}). 
The agent iteratively proposes new programs $\seamprog$ until it meets the stopping criterion.
Lastly, we specify a \textit{seam preferences
skill} (domain-specific system prompt) to condition higher-level direction and strategy for UV seam planning (Sec.~\ref{sec:skill_learning}). This skill contains optional
artist's knowledge that we distilled to help the agent create a higher-quality UV atlas.

\subsection{Queryable mesh representation} 
\label{sec:mesh_db}

We define a preprocessor $\processor$ which, before the agentic loop, computes attributes of the input mesh $\inputmesh$ and stores them in a relational database. This constitutes a queryable representation $\meshdb = \processor(\inputmesh)$ that enables SQL-based queries to retrieve compact mesh information using few LLM tokens, instead of serializing the whole mesh into the \VLM{}'s context.

\mypar{Mesh Attributes} fall into three categories. \textit{Topological attributes} come from connectivity. A mesh is partitioned into topologically connected components, and each face and edge belongs to exactly one component 
\textit{Geometric attributes} are either global (e.g., symmetry planes) or local to an edge or face (position, curvature, normal, length, dihedral angle, thickness, and ambient occlusion). We use ambient occlusion as a proxy for edge visibility, enabling the \VLM{} to query hidden seams. 
\textit{Semantic attributes} are text labels associated with each face and edge; edges between two labels are flagged as semantic boundaries. We obtain them by prompting a \VLM{} to assign labels to mesh partitions defined by a segmentation algorithm~\citep{liu2025partfield}. See Appendix~\ref {app:mesh-schema} for the full list of attributes.

\subsection{Seam-selection DSL} 
\label{sec:selection_dsl}

\begin{figure}
    \centering
    \includegraphics[width=\linewidth]{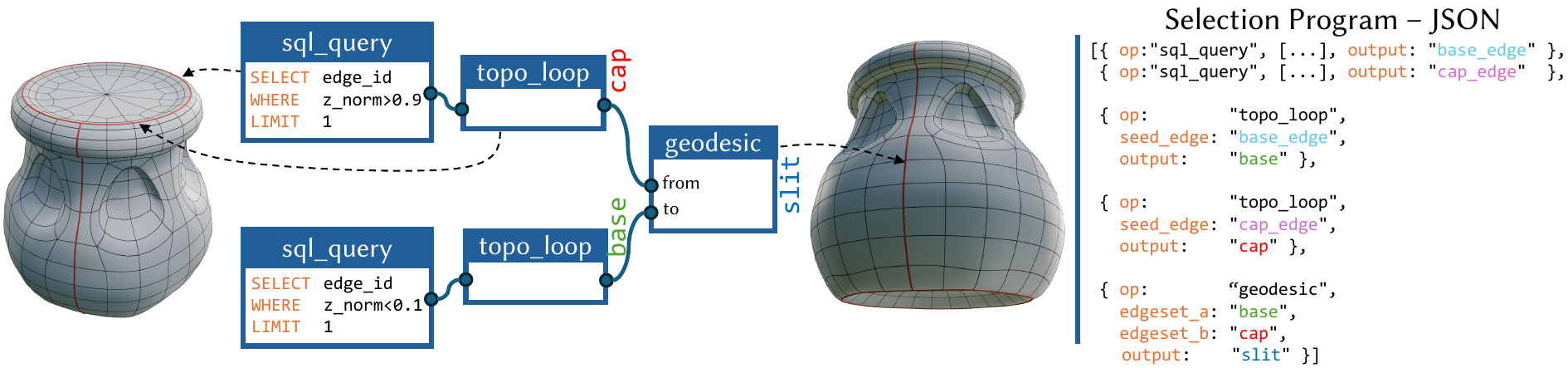}
    \caption{A \textbf{domain-specific language (DSL)} allows expressing edge selection and plan manipulation via a high-level seam selection program (JSON) based on aggregate edge selection operators.}
    \label{fig:dsl}
\end{figure}

A seam plan is a subset of mesh edges, $\seamplan \subseteq E$, selected as seams. The agent specifies this plan through a \emph{seam selection program} $\seamprog$, in a domain-specific language (DSL) whose execution by the interpreter produces $\seamplan$. A program consists of a jointly generated collection of statements 
$\seamprog = \{p_1, \dots, p_N\}.$ When $\seamprog$ is evaluated, a topological sort ensures that no statement is executed before its inputs are available. 
A program is thus rejected if it does not constitute a directed acyclic graph. 
We also statically check that each input has the expected type; otherwise, we reject the program.
Lastly, a program can be rejected at runtime, as some operators may raise exceptions when their result is likely to lead to bad seams. 
Fig.~\ref{fig:dsl} shows our program-based representation of edge sets.

\mypar{Statements} correspond to calls to our built-in edge-selection operators. Each statement $p_i$ specifies an operator $\tau$, a list of input arguments provided to this operator, and an output name. Each input argument is either a constant literal (e.g., an SQL query) or the output of a previous statement, referenced by its name. To assist the agent in exploring the space of possible programs, static typing ensures that the input arguments match the data type operator $\tau$ expects.
Since many 3D shapes exhibit a symmetry along an axis, each statement may carry a \inlinecode{mirror} tag, in which case its inputs are reflected across the detected symmetry plane. In order to ease agent iterations, our DSL also allows the insertion of comments about the intent of a statement.

\mypar{Operators} $\tau$ are the core routines that operate on the mesh representation $\meshdb$ and return sets of edges. Crucially, they enable the \VLM{} to work at a higher level of abstraction than individual edge selection. For example, \inlinecode{topological\_loop} and \inlinecode{geodesic} are operators
that return aggregate sets of edges. Operator arguments may refer to geometric quantities that are difficult for the
\VLM{} to specify exactly; therefore, each operator can be evaluated in \emph{preview mode $\tau^*$}. The \VLM{} previews operators to validate that they produce a non-empty set of edges.  To support this further, the DSL also includes \emph{search} variants $\hat{\tau}^*$. 
Rather than requiring an exact argument $\theta$, a search statement specifies a small candidate region $\theta+\epsilon$, and returns ranked candidates as executable DSL statements (e.g., \inlinecode{find\_loops$(\theta)$} returns ranked candidates \inlinecode{topological\_loop($\theta+\epsilon$)}).  
See Appendix~\ref{tab:operator-complexity} for implementation details of the full operators collection.


\mypar{Interpreter} evaluates the submitted program $\seamprog$ into a seam plan $\seamplan$ expressed as an explicit list of edge indices. It parses all statements $p_i$ and orders them with respect to their dependencies (a statement can only be executed once all its arguments that reference other statements have been evaluated). The interpreter may reject DSL statements and provide informative error messages for cyclic dependencies, static typing issues, or runtime errors.

\subsection{UV quality feedback loop}
\label{sec:submit}

At iteration $\iteration$, the agent evaluates its seam plan $\seamplan_\iteration$ by calling the parameterization algorithm $\Psi$, which flattens the 3D mesh into a 2D UV atlas based on $\seamplan_\iteration$. The agent can then assess its quality using feedback metrics $\metrics_\iteration$, assign an ID to $\seamplan_\iteration$ for future reference (e.g., comparison, rollback), stop iterating once it satisfies the stopping criterion, and return its best plan $\seamplan_\iteration^\star$.


\mypar{Feedback metrics} $\metrics$ help the agent assess the quality of the UV atlas.
resulting from its proposed seam plan $\seamplan_\iteration$ and refine it for the next iteration $\seamplan_{\iteration+1}$. Feedback is typically provided as compact text that consumes few \VLM{} tokens. It features \textit{global UV atlas metrics} such as seam length, number of UV charts, distortion, and seam network connectivity.
It also includes \textit{local UV chart metrics} such as chart convexity and flags seams associated with high-distortion UV charts. In complex situations, the \VLM{} may also request distortion heatmap visualizations rendered on the 3D model. The full implementation of the metrics is listed in Appendix~\ref{app:evaluation-metrics}.

\mypar{Stopping criterion.}
The agent loop terminates when the quality criteria are satisfied or when the maximum time budget is exceeded. The quality criteria require that no face collapses to zero UV area, and that the maximum per-face area distortion and mean conformal distortion are below user-specified thresholds: $D^{\text{max}}_{\text{area}} < c_1 \wedge D_{\text{conf}} < c_2$.

\mypar{Plan selection.} The agent returns the best-scored plan $\seamplan_\iteration^\star$ based on a lexicographic ranking. Plans are first ranked by
feasibility (area and conformal distortion under predefined target thresholds), then by fewer sliver charts (charts below $0.5\%$ of the surface), then
by lower overall and conformal distortion, then by higher convexity and UV space utilization. If no plan is feasible, it returns the one with the smallest relative violation of the distortion thresholds. Note that the preferences skill described below can be used to steer this design towards other specific goals.

\subsection{Seam Preferences Skill}
\label{sec:skill_learning}
To guide the \VLM{} in solving the seam-planning task, we provide API documentation for the agent tools and the seam-selection DSL grammar, with examples. In addition to conditioning the \VLM{} planning strategy, we manually distill seam placement handbooks from publicly available online resources such as \cite{blender_smart_uv} and Maya \cite{autodesk2023uvmapping}. For example, we include recommendations for handling common shapes, as well as situational cases such as head shapes, which, for instance, require a single seam line from the neck to the top of the head.

\section{Experiments}
\label{sec:experiments}

\subsection{Experimental Setup}
\label{sec:experimental-setup}

\mypar{Datasets.}
We evaluate on two 50-asset datasets. We randomly sample 50 assets from 18
available classes of Adobe Substance 3D Assets library~\cite{adobe_substance3d_assets} to form \textsc{Substance-50}.
It contains production-grade quad meshes with artist UVs and spans
1,344--53,634 polygons (median 9,392). In addition, we randomly sample 50 quad meshes from
the Toys4K dataset~\cite{stojanov2021using} to form \textsc{Toys4K-50}, spanning 61--330,640 polygons
(median 4,752). See Appendix~\ref{app:eval-dataset}.

\mypar{Baselines.}
We compare against multiple baselines on the produced seams and related UV atlas on both datasets.
UV atlas receives the same rigid average island scaling and packing via Blender at a 0.001 margin. 
We compare against FlattenAnything~\citep{zhang2024flatten}, OptCuts~\citep{li2018optcuts}, \textsc{xatlas}~\citep{xatlas}, Blender Smart UV~\cite{blender_smart_uv}, and \textsc{PartUV}~\citep{wang2025partuv}. On \textsc{Substance-50}, we also compare with artists' UVs. See Appendix~\ref{app:baselines}.

\mypar{Evaluation protocol.}
We quantify seam and UV atlas quality with several metrics. 
We measure \textit{UV chart count}, \textit{UV seam length} as the total chart-boundary length in UV space, and normalized \textit{3D
seam length}.
To assess seam visibility, we measure mean normalized \textit{seam ambient occlusion (AO)}. We also include 
\textit{Non-Disk} as the fraction of UV charts classified as non-disks,
\textit{Compactness} as a chart's similarity to a perfect circle, chart \textit{Convexity}, and seam 
\textit{Jaggedness} of UV charts' boundary regularity.
Lastly, we compute standard distortion metrics such as \textit{area}, \textit{conformal}, and \textit{texel spread}, and report the metric implementations in the Appendix~\ref{app:evaluation-metrics}.

\mypar{Implementation details.}
We set the \VLM{} to Claude Opus 5
with high reasoning effort, default sampling configuration, and a 1M token context. It operates our tools through the Claude Code Harness. We evaluate the parameterization and DSL operators in Blender 5.1.2 on CPU (see Appendix~\ref{app:parameterization}).
We use PartField as semantic edge- and face- priors.
We set runtime timeout to  $40$ min, and the stopping criterion to distortion constants to  $\big[D^{\max}{\text{area}} < 2.0\big] \land \big[D{\text{conf}} < 1.5\big]$. 

\subsection{Results}
\label{sec:main-results}
\label{sec:exp-production}

\mypar{Quantitative results.} We report the comprehensive benchmark in Table~\ref{tab:uv-benchmark}. Our method outperforms all competitor baselines on seam-quality metrics and produces a valid seam set and UV atlas for all mesh sizes. Compared to the most competitive baseline, our method achieves 2.90×/4.29× fewer UV charts, 1.63×/1.69× shorter UV seams, and 1.44×/1.66× shorter 3D seams than \textsc{PartUV} on \textsc{Substance-50/Toys4k-50} with minimal distortion cost.
Furthermore, our metric statistics closely align with the artist reference; e.g., 19.6 for ours vs. 19.5 in the number of UV charts. Our method consistently produces high-quality seams that yield clean cuts on the 3D surface, especially in more occluded areas whenever possible.
However, seam placement is constrained by limitations of the UV parameterization. 
In production, even charts cut along artist-authored seams are typically refined manually by relaxing and pinning UV coordinates before texturing. This refinement step is orthogonal to seam generation and therefore outside the scope of this work.

\mypar{Human evaluation results.} We conduct an A/B study comparing Ours against \textsc{PartUV}, the strongest baseline in Tab.~\ref{tab:uv-benchmark}.
From a fixed randomly sampled pool of 80 meshes (40 from each evaluation set), every annotator rates a random subset of 40 (20 per set), choosing the better seams or marking a tie.
Across 15 annotators (10 professional artists) and 600 ratings, annotators prefer Ours on 79.7\% of decisive ratings  (ties excluded); the preference holds on both \textsc{Substance-50} 75.7\% and \textsc{Toys4K50} 83.6\%.
Restricted to the 10 professional artists, the  overall preference  (ties excluded) is 80.9\%.
Beyond the pairwise comparisons, annotators also flag whether the preferred seams are usable as a hot-start for production work. Ours-preferred seams are marked at 53.0\%.
Our approach therefore aligns more closely with professional artist expectations than the strongest available baseline.

\begin{figure*}[t]
  \centering
  \includegraphics[width=\textwidth]{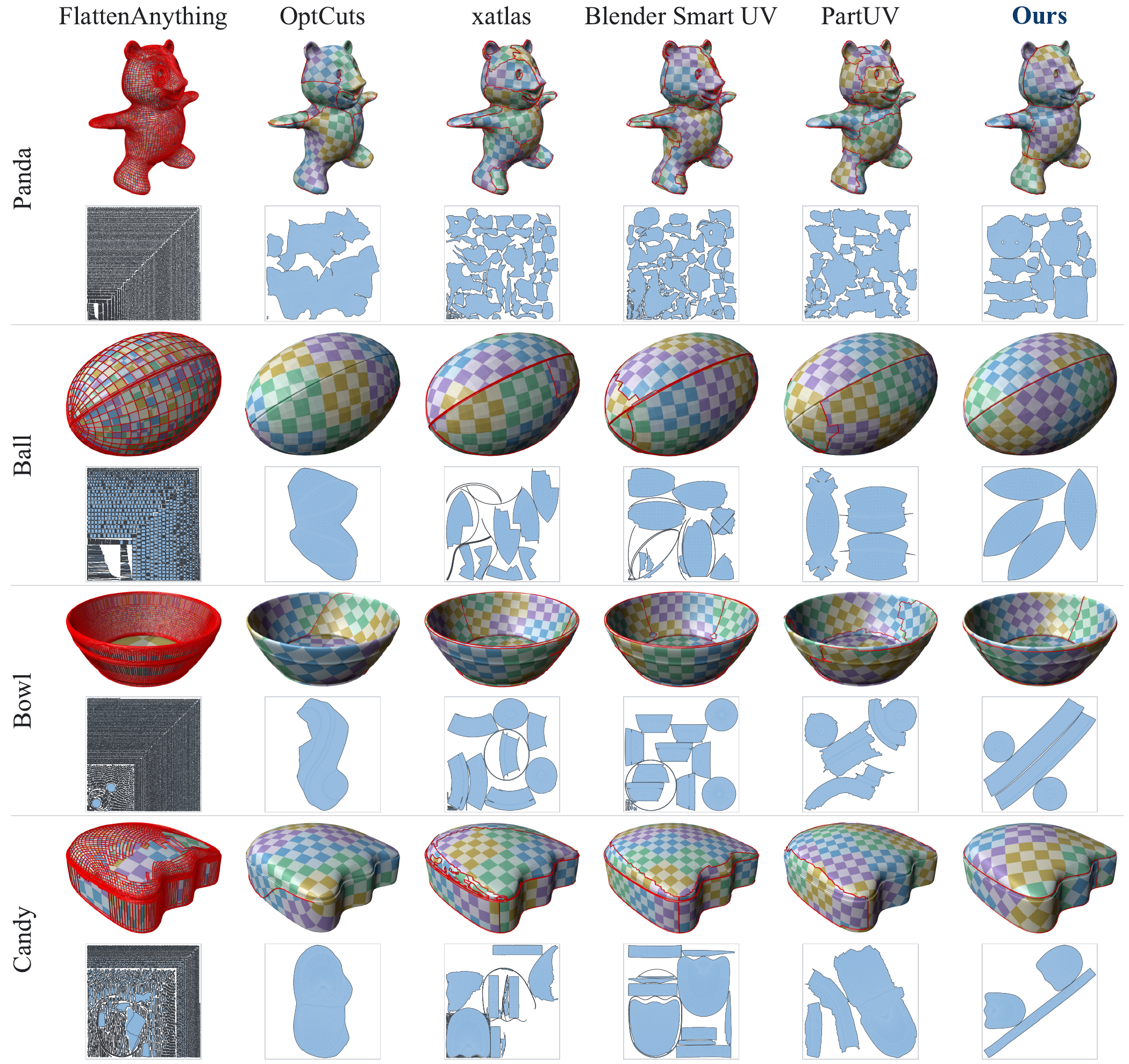}
  \caption{\textbf{Qualitative comparison on \textsc{Toys4K-50}.} For each
  asset, the top row shows a UV checkerboard and generated seams in red, while the bottom row shows the parametrized UV atlas.}
  \label{fig:qualitative}
\end{figure*}

\mypar{Qualitative results.}
Fig.~\ref{fig:qualitative} shows  comparisons to baselines on the \textsc{Toys-4k} dataset. \method{} solves the UV unwrapping through a minimal set of seams.
Generally, our method produces seams in occluded areas, e.g., under the Panda's arms, or around high-dihedral-angle edges, e.g., Candy, and follows the natural 
structure of typical objects when mesh attributes such as ambient occlusion and curvature are at parity, e.g., the Ball.

\begin{table*}[t]
  \centering
  \caption{\textbf{UV parameterization benchmark}. Here, $(n)$ is the number of successful runs. We highlight in \textbf{bold} and \underline{underlined}
  values that indicate the best and second-best results, respectively. Methods producing failure cases (\textsuperscript{\textdagger}) or excessively high mean distortion (\textsuperscript{\textdaggerdbl}) are excluded from the ranking.}
  \label{tab:uv-benchmark}
  \begingroup
  \setlength{\tabcolsep}{3.1pt}
  \renewcommand{\arraystretch}{1.14}
  \scriptsize
  \resizebox{\textwidth}{!}{%
  \begin{tabular}{@{}lccccccccccc@{}}
    \toprule
    & \multicolumn{8}{c}{Seam quality}
    & \multicolumn{3}{c}{Distortion} \\
    \cmidrule(lr){2-9}\cmidrule(lr){10-12}
    Method ($n$)
    & \shortstack{UV\\Charts $\downarrow$}
    & \shortstack{UV Seam\\length $\downarrow$}
    & \shortstack{3D Seam\\length $\downarrow$}
    & AO $\uparrow$
    & Non-disk $\downarrow$
    & Compact. $\uparrow$
    & Convex. $\uparrow$
    & Jagged. $\downarrow$
    & \shortstack{Texel\\spread $\downarrow$}
    & \shortstack{Area\\dist. $\downarrow$}
    & \shortstack{Conformal\\dist. $\downarrow$} \\
    \midrule
    \multicolumn{12}{@{}l}{\textbf{Substance-50}} \\
    \addlinespace[2pt]
    FlattenAnything~\cite{zhang2024flatten}\textsuperscript{\textdaggerdbl} (50) & (19.7) & (17.0) & (5.9) & (0.545) & (0.298) & (3.042) & (0.902) & (0.58) & (1.703) & (4.067) & (43.716) \\
    OptCuts~\cite{li2018optcuts}\textsuperscript{\textdagger} (38) & (3.8) & (6.3) & (3.4) & (0.504) & (0.051) & (0.473) & (0.819) & (0.84) & (0.197) & (1.172) & (1.133) \\
    \textsc{xatlas}(LSCM)~\cite{levy2002} (50) & 355.5 & 44.2 & 26.6 & 0.503 & 0.065 & 0.428 & \underline{0.853} & 1.21 & \underline{0.163} & \textbf{1.095} & 1.111 \\
    Blender Smart UV~\cite{sheffer2006survey} (50) & 162.6 & 37.9 & 21.5 & \underline{0.521} & 0.071 & 0.408 & 0.807 & 1.14 & \textbf{0.159} & \underline{1.102} & 1.162 \\
    \textsc{PartUV}~\cite{wang2025partuv} (50) & \underline{56.8} & \underline{20.6} & \underline{9.8} & 0.493 & \underline{0.055} & \underline{0.444} & 0.792 & \textbf{0.74} & 0.171 & 1.117 & \textbf{1.030} \\
    \rowcolor{BenchmarkBlueLight}
    \textbf{\textcolor{BenchmarkBlue}{Ours}} (50) & \textbf{19.6} & \textbf{12.6} & \textbf{6.8} & \textbf{0.527} & \textbf{0.014} & \textbf{0.611} & \textbf{0.891} & \underline{0.78} & 0.179 & 1.178 & \underline{1.101} \\
    \rowcolor{BenchmarkGrayLight}
    Artist reference (50) & 19.5 & 13.8 & 7.0 & 0.547 & 0.059 & 0.568 & 0.886 & 0.50 & 0.166 & 1.117 & 1.086 \\
    \midrule
    \multicolumn{12}{@{}l}{\textbf{Toys4K-50}} \\
    \addlinespace[2pt]
    FlattenAnything~\cite{zhang2024flatten}\textsuperscript{\textdagger} (49) & (13974.4) & (108.1) & (159.2) & (0.499) & (0.002) & (0.509) & (0.984) & (0.91) & (0.761) & (1.596) & (17.187) \\
    OptCuts~\cite{li2018optcuts}\textsuperscript{\textdagger} (37) & (1.5) & (4.6) & (3.7) & (0.490) & (0.198) & (0.413) & (0.786) & (0.53) & (0.190) & (1.183) & (1.140) \\
    \textsc{xatlas} (LSCM)~\cite{levy2002} (50) & 885.0 & 33.6 & 24.6 & 0.519 & \textbf{0.040} & 0.418 & \underline{0.841} & \underline{0.68} & \textbf{0.138} & \textbf{1.094} & 1.386 \\
    Blender Smart UV~\cite{sheffer2006survey} (50) & 151.6 & 23.9 & 17.3 & \underline{0.522} & 0.067 & 0.394 & 0.812 & 0.70 & \underline{0.156} & \underline{1.097} & \underline{1.156} \\
    \textsc{PartUV}~\cite{wang2025partuv} (50) & \underline{39.9} & \underline{15.0} & \underline{10.1} & 0.501 & 0.108 & \underline{0.425} & 0.804 & 0.82 & 0.192 & 1.143 & \textbf{1.054} \\
    \rowcolor{BenchmarkBlueLight}
    \textbf{\textcolor{BenchmarkBlue}{Ours}} (50) & \textbf{9.3} & \textbf{8.9} & \textbf{6.1} & \textbf{0.533} & \underline{0.063} & \textbf{0.610} & \textbf{0.904} & \textbf{0.47} & 0.215 & 1.201 & 1.172 \\
    \bottomrule
  \end{tabular}%
  }
  \endgroup
\end{table*}

\begin{figure*}[t]
  \centering
  \includegraphics[width=\textwidth]{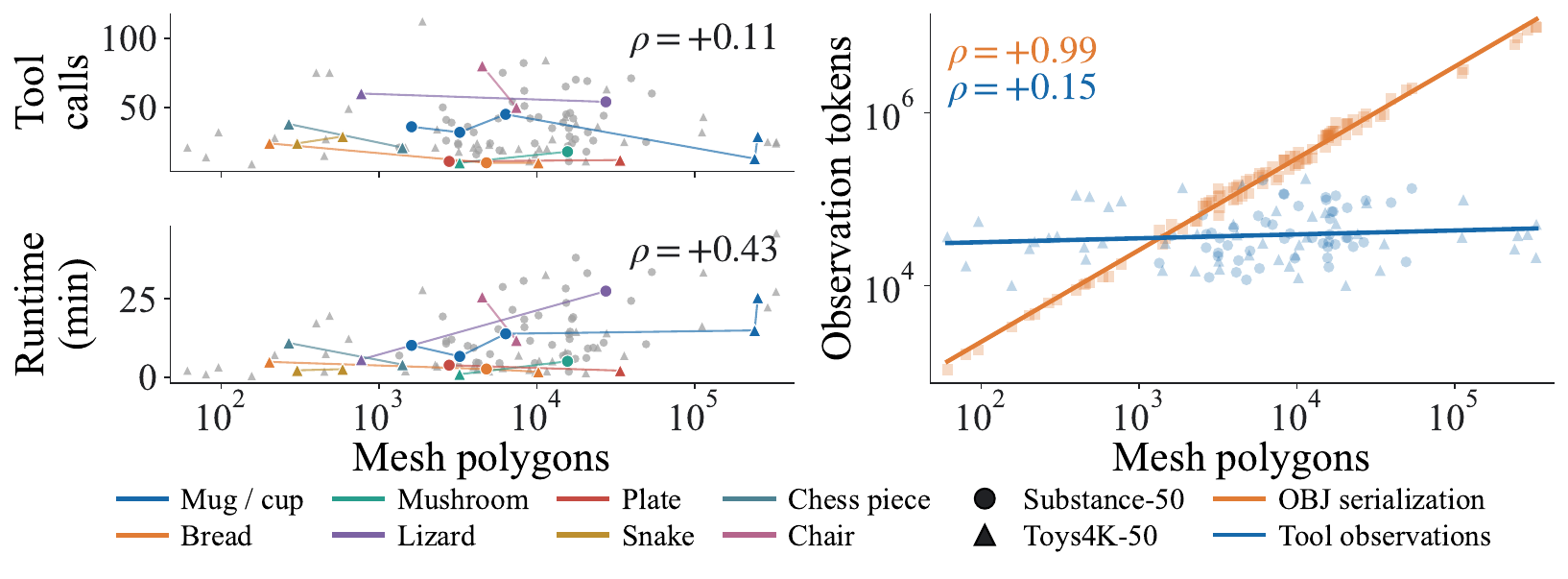}
  \caption{\textbf{Tool calls, runtime, and token consumption across varying mesh polygon sizes.}
  \textit{Left}: 
Colored paths connect meshes from the same semantic class, but with different geometry and topology.
Tool call exhibits high-level agent reasoning, has negligible correlation with mesh size, while runtime scales super-linearly due to parameterization feedback cost per iteration. \textit{Right}: \VLM{} token cost across varying mesh polygon sizes is near-constant, while direct mesh serialization grows linearly.}
  \label{fig:resolution-scaling}
\end{figure*}

\mypar{Mesh size performance scaling.}  In Fig.~\ref{fig:resolution-scaling}(left), we show dot plots of tool calls and wall-clock runtime across increasing polygon sizes.
Notably, the number of tool calls (top left), which reflects the agent's high-level reasoning, correlates negligibly with mesh size (Spearman $\rho=+0.11$ over all meshes). For meshes with similar semantics, geometry, and topology, it interpolates toward a constant value. 
We argue that this level of abstraction enables reasoning to depend on shape complexity rather than polygon count. For example, mugs - despite having a high vertex count range of $1,608-250,296$ (157x span)- and Plate at $2,786-33,794 $ (12x span) - retain similar high-level plans and tool-call counts that lead to high-quality UV parameterizations. 
While the operators are bounded by $O(n \log n)$ time complexity as the polygon count $n$ increases, the comprehensive solution runtime (bottom left), however, is bounded by the feedback loop parameterizations that require sparse linear solves over \(O(n)\) mesh vertices, yielding superlinear time \(O(k n^{3/2})\) for manifold meshes under sparse direct factorization.
Our experiments achieve a 11.8-minute runtime, of which 71\% is model reasoning and about 29\% is tool server time. \method{} scales up to $300k$ polygons, one order of magnitude higher than autoregressive methods~\cite{li2025autoregressive, xu2026seamgen}.

\mypar{Mesh Preprocessing Time.} Building the mesh representation $\meshdb$ is a one-off cost which takes median $2.6$ minutes (range $1.7$--$15.0$), of
which the relational representation is $2.8$ seconds (at most $21.3$ seconds on the largest asset). The majority of the cost comes from PartField segmentation over a fixed number of partitions ($17.4\%$, median $36.8$
seconds) and a one-time \VLM{} labeling cost per partition  (totaling $77.9\%$, median $105.8$ seconds). See Appendix Fig.~\ref{fig:vlm-labelling}.

%
\mypar{\VLM{} Token cost.}  The model reads a near-constant amount of text tokens about a mesh regardless of its size
(Fig.~\ref{fig:resolution-scaling}(right)). Over $100$ runs, the tool output read by the agent has a
median of $35,934$ tokens per run (Spearman $\rho=+0.15$ against $\log$ polygon count), and the outer
face quartiles differ by only $1.12\times$; the budget is similar on both datasets
(\textsc{Substance-50} median 35,924, \textsc{Toys4K-50} 36,159). Most of this text is feedback from the parametrized UV atlas (67\%) rather than perception (25\%). 
Conversely,  serializing the mesh instead would scale linearly with resolution: the same asset's OBJ file is
a median of 229,604 tokens ($\rho=+0.99$), exceeding the observed text by up to $465\times$ on the largest meshes. 
This compact representation results in a median context size of 157k tokens (max 503k, $\rho=+0.23$). 
Model output is dominated by thinking (median 38,922 of 62,940 output tokens per run)
and tracks asset complexity rather than polygon count.





\subsection{Ablation Studies}
\label{sec:ablations}

\begin{table}[t]
  \centering
  \caption{\textbf{Ablations.} We ablate the feedback loop (a) and the LLM model (b) and report distortion metrics (red indicates regression).}
  \label{tab:ablations}
  \begingroup
  \footnotesize
  \definecolor{AblationGain}{RGB}{0,110,65}
  \definecolor{AblationLoss}{RGB}{180,45,35}
  \newcommand{\AblRef}[1]{#1}
  \newcommand{\AblGain}[2]{#1\,{\tiny\textcolor{AblationGain}{$#2$}}}
  \newcommand{\AblLoss}[2]{#1\,{\tiny\textcolor{AblationLoss}{$#2$}}}
  \setlength{\tabcolsep}{4pt}
  \renewcommand{\arraystretch}{1.0}
  \begin{tabular*}{\linewidth}{@{\extracolsep{\fill}}llllll@{}}
    \toprule
    \multicolumn{2}{@{}l}{Variant} & Area dist. $\downarrow$ & Angular dist. $\uparrow$
            & Conformal dist. $\downarrow$ & Texel spread $\downarrow$ \\
    \midrule
    \multicolumn{6}{@{}l}{\textbf{(a) Closed-loop feedback}} \\
Substance-50 & \textbf{With feedback (ref.)} & \AblRef{1.151} & \AblRef{0.928} & \AblRef{1.100} & \AblRef{0.178} \\
 & No closed loop & \AblLoss{1.341}{+16.5\%} & \AblLoss{0.887}{-4.4\%} & \AblLoss{1.419}{+29.0\%} & \AblLoss{0.377}{+111.6\%} \\
    \addlinespace[2pt]
Toys4K-50 & \textbf{With feedback (ref.)} & \AblRef{1.201} & \AblRef{0.903} & \AblRef{1.173} & \AblRef{0.215} \\
 & No closed loop & \AblLoss{1.525}{+27.0\%} & \AblLoss{0.822}{-8.9\%} & \AblLoss{1.359}{+15.9\%} & \AblLoss{0.512}{+137.8\%} \\
    \midrule
    \multicolumn{6}{@{}l}{\textbf{(b) LLM model}} \\
Substance-50 & \textbf{Claude Opus 5 (ref.)} & \AblRef{1.151} & \AblRef{0.928} & \AblRef{1.101} & \AblRef{0.178} \\
 & Qwen3.8-27B & \AblLoss{1.362}{+18.3\%} & \AblLoss{0.854}{-7.9\%} & \AblLoss{1.370}{+24.4\%} & \AblLoss{0.348}{+95.4\%} \\
    \addlinespace[2pt]
Toys4K-50 & \textbf{Claude Opus 5 (ref.)} & \AblRef{1.201} & \AblRef{0.903} & \AblRef{1.172} & \AblRef{0.215} \\
 & Qwen3.8-27B & \AblLoss{1.450}{+20.7\%} & \AblLoss{0.838}{-7.2\%} & \AblLoss{1.750}{+49.4\%} & \AblLoss{0.390}{+81.1\%} \\
    \bottomrule
  \end{tabular*}
  \endgroup
\end{table}

%

\begin{figure*}[t]
  \centering
  \includegraphics[width=\textwidth]{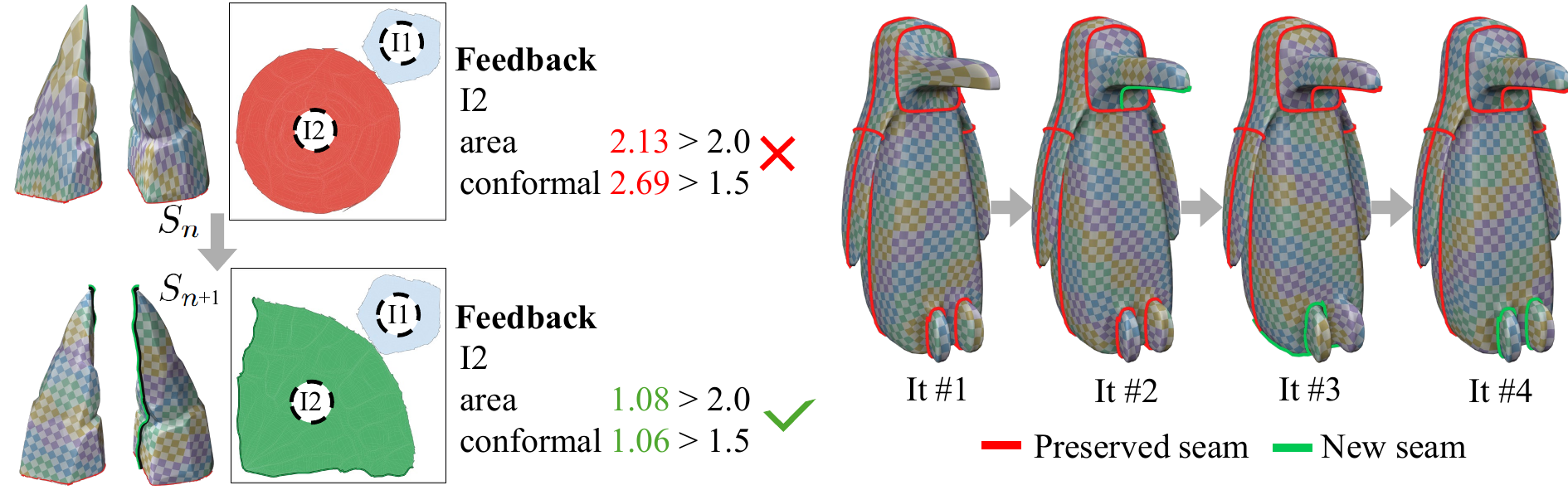}
  \caption{\textbf{Closed-loop feedback.
    \textit{(Left)}} The cone of the rock chart fails distortion bounds. After refinement, the agent adds one apex-to-base slit, opening the cone into a fan while leaving the base chart untouched. \textit{Right}: a penguin model with the generated seams at each iteration.}
  \label{fig:feedback-evolution}
\end{figure*}

\mypar{Feedback Loop.} In Tab~\ref{tab:ablations}
(a),
we observe that the plans with feedback loops are better than the first attempt on every distortion metric.   
The UV feedback raises the share of meshes inside the acceptable distortion gate from 66\% on the first attempt to 97\%. The agent applied 342 plan versions
over the 100 meshes, a median of 3 per mesh (mean 3.42, max 9).
Most gains occurred within 3 iterations, reaching 91 accepted plans, and after 9 revisions, reaching 97\% acceptance.
Therefore, we consider the closed-loop feedback as an essential component of our solution. See Fig.~\ref{fig:feedback-evolution} for an example.

\mypar{LLM model comparison.} In Tab~\ref{tab:ablations}
(b)
we compare benchmark evaluations between Claude Opus 5 and Qwen3.8-27B.  We see a significant improvement when using frontier models with stronger reasoning and higher token capacity (200K vs 1M context tokens).  Nonetheless, the small open-source model completed the task, though at higher distortion costs, validating that the proposed method is portable to smaller models.

\section{Conclusion} 
\label{sec:conclusion}
We introduced \method{}, an agentic framework for UV unwrapping in which \aVLM{} perceives a mesh through compact queries, acts via a seam-selection DSL, and refines its plan based on parameterization feedback. We let the agent operate at a higher level of abstraction than individual edge selection, via aggregate operators, enabling the \VLM{} to focus on high-level intent.
Our experimental results show strong human alignment, achieving over 80.9\% professional artist preference over SOTA baselines across a benchmark of small to large meshes and hard and organic surfaces.
In addition, artists rate 53\% of our meshes as usable as hot-starts for production use.
Lastly, our mesh and DSL representation is highly compact and token-efficient, enabling \aVLM{} to efficiently reason on large and complex meshes with $10^5$ faces, an order of magnitude higher than autoregressive seam placement methods.

\mypar{Challenges and future work.} We focus on quad meshes, as they are well suited for existing multi-edge selection tools, like edge loops and are the main class used in production.
Our main contributions (DSL, mesh queries, skill files) theoretically generalize to triangular meshes and early explorations with SeamCut~\cite{lucquin2017seamcut} are encouraging. Regarding runtime, there are two bottlenecks. First, PartField~\cite{liu2025partfield} can be costly for larger meshes. We expect that improvements in 3D segmentation will transfer directly to speed and accuracy improvements for \method{}. Second, the iterative agentic loop is expensive. We expect that efficiency gains from general agentic frameworks and VLMs (e.g., Qwen in \autoref{tab:ablations}) as well as more informative evaluation signals would translate in fewer cycles. Finally, we focus on the placement of seams. In future work, we want to explore the interaction of parameterization jointly with UV-placement, e.g., for refining UV-maps or selecting the right parameterization approach for a given topological disk.

\bibliography{iclr2027_conference}
\bibliographystyle{iclr2027_conference}
\clearpage
\appendix
\suppressfloats[t]

\begin{center}
  {\LARGE Appendix}
\end{center}

\setlength{\textfloatsep}{10pt plus 2pt minus 2pt}
\setlength{\floatsep}{8pt plus 2pt minus 2pt}
\setlength{\intextsep}{10pt plus 2pt minus 2pt}

\section{Implementation Details}
\label{app:implementation_details}

\subsection{Datasets}
\label{app:eval-dataset}

\begin{figure}[t]
    \includegraphics[width=1\linewidth]{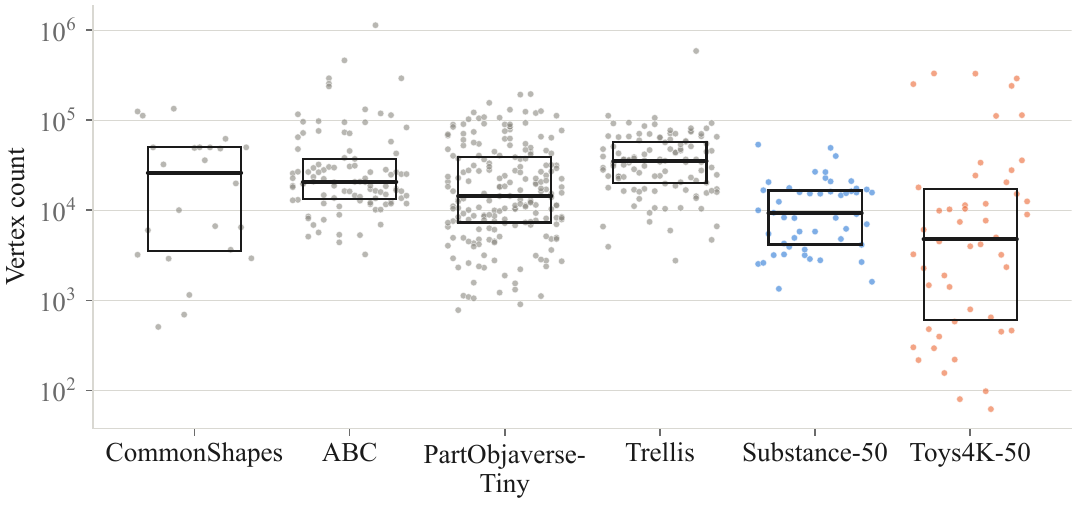}
    \caption{\textbf{Dataset mesh size comparison.} We show the overall distribution of mesh sizes across common 3D mesh datasets\cite{jacobson2013common3dtestmodels,koch2019abc,deitke2023objaverse,yang2024sampart3d,xiang2025structured}. Substance-50 and Toys4k-50 have comparable mesh size distribution with respect to the others, measured by the number of vertices.}
    \label{fig:dataset_comparison}
\end{figure}

\begin{figure}[p]
  \includegraphics[width=\textwidth,height=0.84\textheight,keepaspectratio]{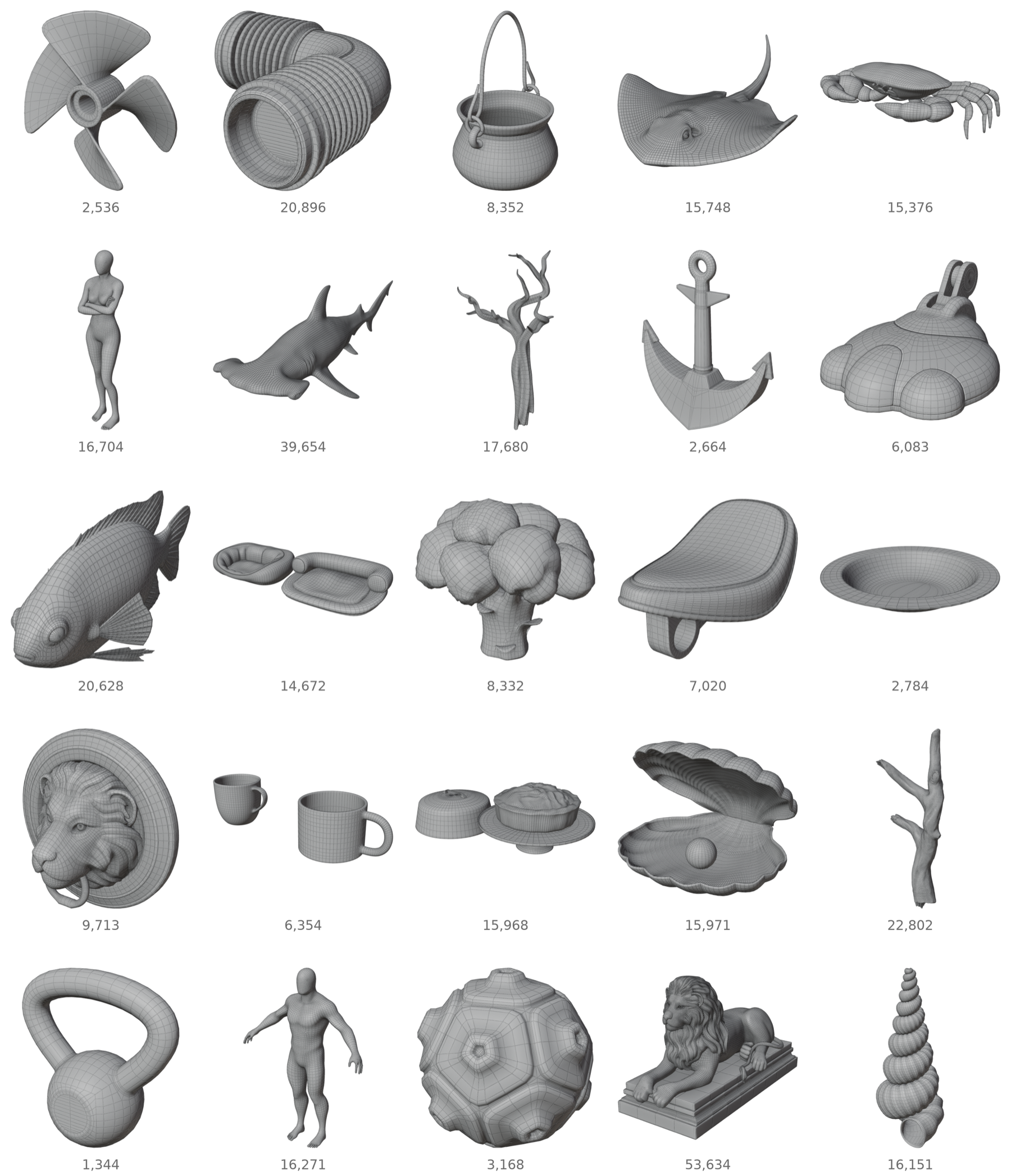}
  \caption{\textsc{Substance-50} evaluation assets.}
  \label{fig:dataset_appendix_1}
\end{figure}

\begin{figure}[p]
  
  \includegraphics[width=\textwidth,height=0.84\textheight,keepaspectratio]{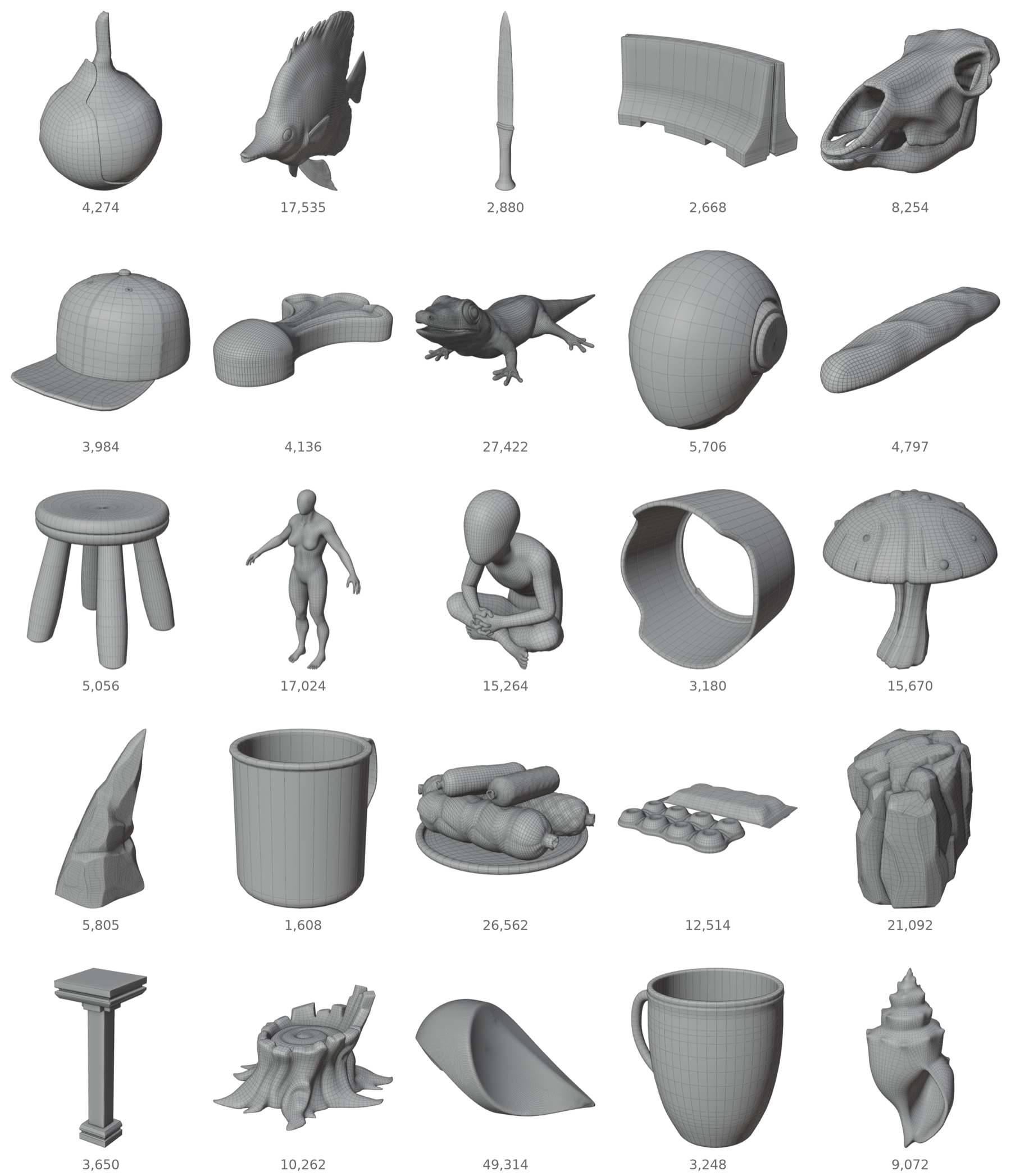}
  \caption{\textsc{Substance-50} evaluation assets (continued).}
  \label{fig:dataset_appendix_2}
\end{figure}

\begin{figure}[p]
  
  \includegraphics[width=\textwidth,height=0.84\textheight,keepaspectratio]{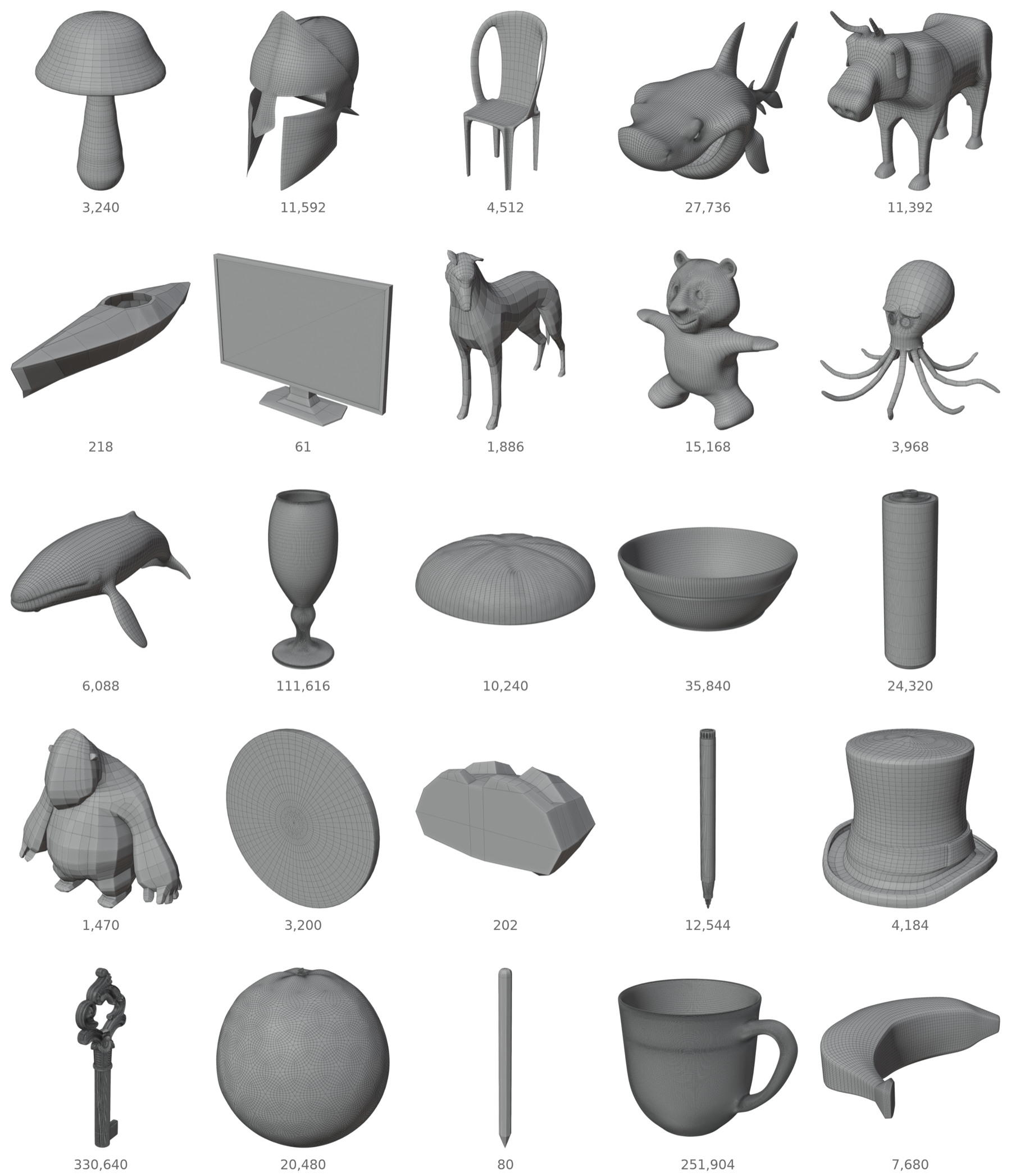}
  \caption{\textsc{Toys4K-50} evaluation assets.}
  \label{fig:toys4k_appendix_1}
\end{figure}

\begin{figure}[p]
  
  \includegraphics[width=\textwidth,height=0.84\textheight,keepaspectratio]{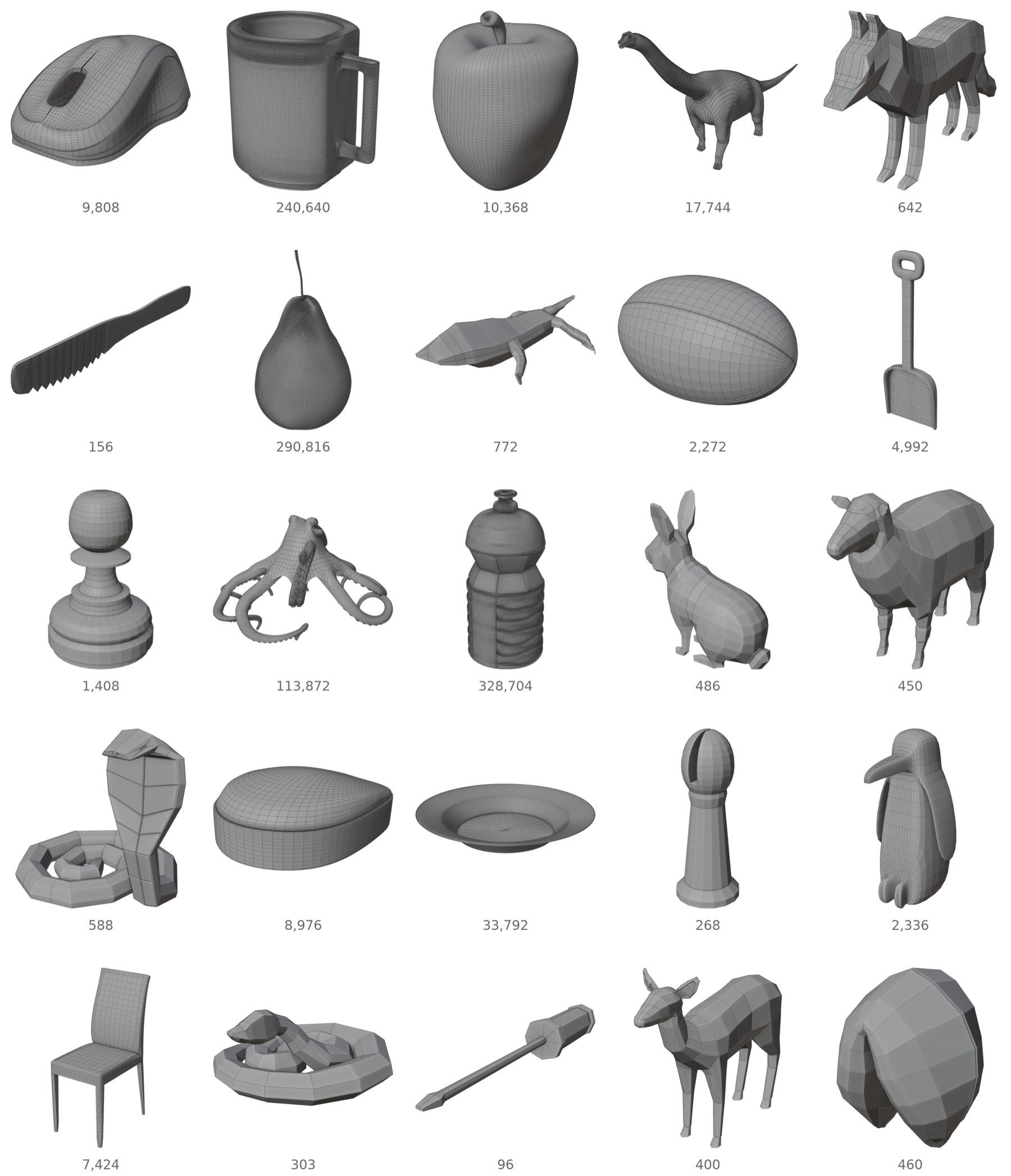}
  \caption{\textsc{Toys4K-50} evaluation assets (continued).}
  \label{fig:toys4k_appendix_2}
\end{figure}

Table~\ref{tab:eval-set-statistics} extends the dataset description in
Sec.~\ref{sec:experimental-setup}. \textsc{Substance-50} spans 18 distinct categories of the Adobe Asset Library, has  28 assets with multiple connected components; 27 are
bilaterally symmetric, and five have an estimated genus of at least two.
\textsc{Toys4K-50} covers 44 categories and contains single-component,
2-manifold meshes; 29 are
bilaterally symmetric, and one has estimated genus at least two.
The distribution of mesh sizes is comparable to commonly used 3D mesh datasets, as shown in Fig.~\ref {fig:dataset_comparison}. In contrast, \textsc{Substance-50} includes a selection of production-grade and manually artist-composed quad meshes. 
Fig.~\ref{fig:dataset_appendix_1}--\ref{fig:toys4k_appendix_2} show all
evaluation assets.

\begin{table}[t]
  
  \caption{\textbf{Dataset statistics.} Entries are
  minimum / median / maximum over their respective set.}
  \label{tab:eval-set-statistics}
  \small
  \setlength{\tabcolsep}{5pt}
  \begin{tabular*}{\linewidth}{@{\extracolsep{\fill}}lcc@{}}
    \toprule
    Property & \textsc{Substance-50} & \textsc{Toys4K-50} \\
    \midrule
    Vertices & 1,344 / 9,233 / 53,638 & 62 / 4,751 / 330,654 \\
    Object parts & 1 / 1 / 6 & 1 / 1 / 1 \\
    Connected components & 1 / 2 / 31 & 1 / 1 / 1 \\
    Triangle fraction & 0 / 0.003 / 0.095 & 0 / 0 / 0.050 \\
    \bottomrule
  \end{tabular*}
\end{table}

\subsection{DSL operators}
We illustrate the set of tools that compose the agent interface at Tab~\ref{tab:operator-complexity}.  For every tool, we report its signature,  description, and time complexity. We include a tool for each DSL operator, plus additional tools for mesh observation and plan management.

\definecolor{OpSection}{RGB}{70,120,170}    
\definecolor{OpGroup}{RGB}{214,228,242}     
\definecolor{OpName}{RGB}{18,58,102}        
\definecolor{OpTimeBg}{RGB}{242,245,249}    
\newcommand{\opname}[1]{\textcolor{OpName}{\textbf{\texttt{#1}}}}
\newcommand{\opret}[1]{\newline\hspace*{0.8em}$\rightarrow$ \textcolor{OpName}{\texttt{#1}}}
\newcommand{\opsearch}{\textcolor{OpName}{\textbf{Searches}}}
\newcommand{\opgroup}[1]{\multicolumn{3}{@{}l}{\cellcolor{OpGroup}\textcolor{OpName}{\emph{#1}}} \\}
\newcommand{\opsection}[1]{\multicolumn{3}{@{}l}{\cellcolor{OpSection}\textcolor{white}{\textbf{#1}}} \\}

\begin{table}[p]
  \centering
  \scriptsize
  \setlength{\tabcolsep}{3pt}
  \renewcommand{\arraystretch}{1.05}
  \caption{\textbf{Agent Tool interface.} ``?''    marks an optional argument; ``$\rightarrow$'' gives the return type. A
    \texttt{Base} statement is evaluated from SQL selections alone; a
    \texttt{Dependent} statement takes the output of a \texttt{Base} statement
    as an argument, so a program is a two-layer dependency graph with in-degree
    at most one. Time complexity is given in $n$, the face count of the mesh;
    $m$ is the number of statements in the submitted program and $r$ the
    number of iterations of the flattener.}
  \label{tab:operator-complexity}
  \begin{tabularx}{\linewidth}{@{}>{\raggedright\arraybackslash}p{0.29\linewidth} X >{\columncolor{OpTimeBg}\raggedright\arraybackslash}p{0.115\linewidth}@{}}
    \toprule
    \textbf{Tool} & \textbf{Details} & \multicolumn{1}{>{\columncolor{white}}l@{}}{\textbf{Time}} \\
    \midrule
    \opsection{1.\ Perception}
    \opgroup{Query}
    \opname{query}(SQL)\opret{SQL-edge $|$ SQL-face, Stats} & Is the main
      tool of the agent for collecting mesh attribute observations. Returns
      valid SQL-edges or SQL-faces fitting the query, as well as their
      aggregate distribution statistics.
      & $O(n)$ \\
    \addlinespace[2pt]
    \opgroup{DSL statements \texttt{Base}: evaluated from SQL selections}
    \opname{topological\_loop}(SQL-edge)\opret{Base} & from an input
      SQL-Edge, it proceeds through a topological walk over the mesh and
      returns the set of connected edges. the chain that goes straight through
      every valence-4 vertex via the opposite edge, ending at poles or the
      boundary
      & $O(n)$ \\
    \opname{boundary\_loop}(SQL-face)\opret{Base} & returns Mesh boundary
      loop as the edges with exactly one incident face, chained into cycles
      & $O(n)$ \\
    \opname{region\_boundary}(SQL-face)\opret{Base} & Returns the boundary
      loop of the input face set: the edges that have exactly one incident
      face inside the set, chained into a closed or open loop.
      & $O(n)$ \\
    \opname{thin\_loop}(SQL-edge)\opret{Base} & \opsearch{} for a thin,
      two-sided part into its front and back faces and returns the loop of
      edges where the two sides meet. The two sides are found by 33
      breadth-first splits, each separating the faces at a different
      fold-sharpness threshold, plus one vote by the sign of each face normal
      against the sheet's PCA plane. The division with a single loop and the
      most balanced sides is returned.
      & $O(n)$ \\
    \opname{semantic\_interface}(Label, Label)\opret{Base} & Returns the set
      of edges of connected edges whose incident faces carry different
      semantic labels. The sets are chained into connected groups, sorted by
      size.
      & $O(n)$ \\
    \opname{seamcut\_seam}(SQL-edge, SQL-edge, SQL-edge\,\dots, $\beta$?)\opret{Base}
      & Returns a crease-following path through 2--8 ordered waypoints, one
      segment per consecutive pair. Each segment is an anisotropic geodesic
      (SeamCut): a short-term vector Dijkstra distance field with look-back 3
      under a curvature-aligned metric of strength $\beta$, then a streamline
      traced from the target back to the source and snapped onto mesh edges.
      Unlike \texttt{geodesic}, it hugs a curved crease instead of cutting
      straight across.
      & $O(n\log n)$ \\
    \opname{crease\_select}(Angle, SQL-edge?)\opret{Base} & Returns an
      unordered set of edges whose dihedral angle is at least \texttt{Angle},
      in one scan of the edge table. Without an angle, it keeps the sharpest
      decile, dihedral rank $\geq 0.9$.
      & $O(n)$ \\
    \opname{find\_loop}(SQL-edge?)\opret{Base[]} & \opsearch{} an area for
      every distinct \texttt{topological\_loop($\theta$)} testing up to 24
      $\theta$-edges by farthest-point sampling. Loops are ranked by coverage,
      then flatness, and duplicates on the same plane are dropped.
      & $O(n)$ \\
    \opname{find\_extremities}(SQL-edge)\opret{Base[]} & \opsearch{} each a
      component for and returns a \texttt{region\_boundary($\theta$)} end.
      Protrusions are the local maxima of surface distance from the component
      center, found by the Dijkstra Algorithm.
      & $O(n\log n)$ \\
    \addlinespace[2pt]
    \opgroup{DSL statements \texttt{Dependent}: attached to \texttt{Base} outputs}
    \opname{geodesic}(Base\,$|$\,SQL-edge, Base\,$|$\,SQL-edge, SQL-edge\,\dots)\opret{Dependent}
      & Returns the shortest chain of edges from a start edge to an end edge,
      Via A$^{*}$ search over the edge-adjacency graph, guided by
      straight-line distance and penalizing turns. A \texttt{Ref} points to
      the realized edges of the statement and snaps to one of those edges.
      Both anchors must lie on the same component.
      & $O(n\log n)$ \\
    \midrule
    \opsection{2.\ Validation}
    \opname{submit\_plan}(program)\opret{PlanVersion} & Interprets a program
      $P$. \texttt{Base} statements are realized first, \texttt{Dependent}
      afterward, mirror twins last. A refinement stage snaps jagged loops,
      drops duplicate seams, and merges rings one face apart or thin parallel
      bands. A cut-topology certificate then checks that every chart is a
      disc, genus 0 with at least one boundary.
      & $O(m\,n\log n)$ \\
    \opname{apply\_seams}()\opret{Atlas, Feedback} & First parametrizes the
      seams S (options SLIM; LSCM, ABF++), then packs UV charts, and
      evaluates
      & $O(r\,n^{3/2})$ \\
    \opname{inspect\_cut\_topology}()\opret{CutTopology} & Computes the cut
      topology of a plan: the face regions the seams separate, the connected
      seam networks, and free seam ends that do not join another seam.
      & $O(n\log n)$ \\
    \bottomrule
  \end{tabularx}
\end{table}

\subsection{Mesh Representation}
\label{app:mesh-schema}

\begin{figure}[t]
  \includegraphics[width=\linewidth]{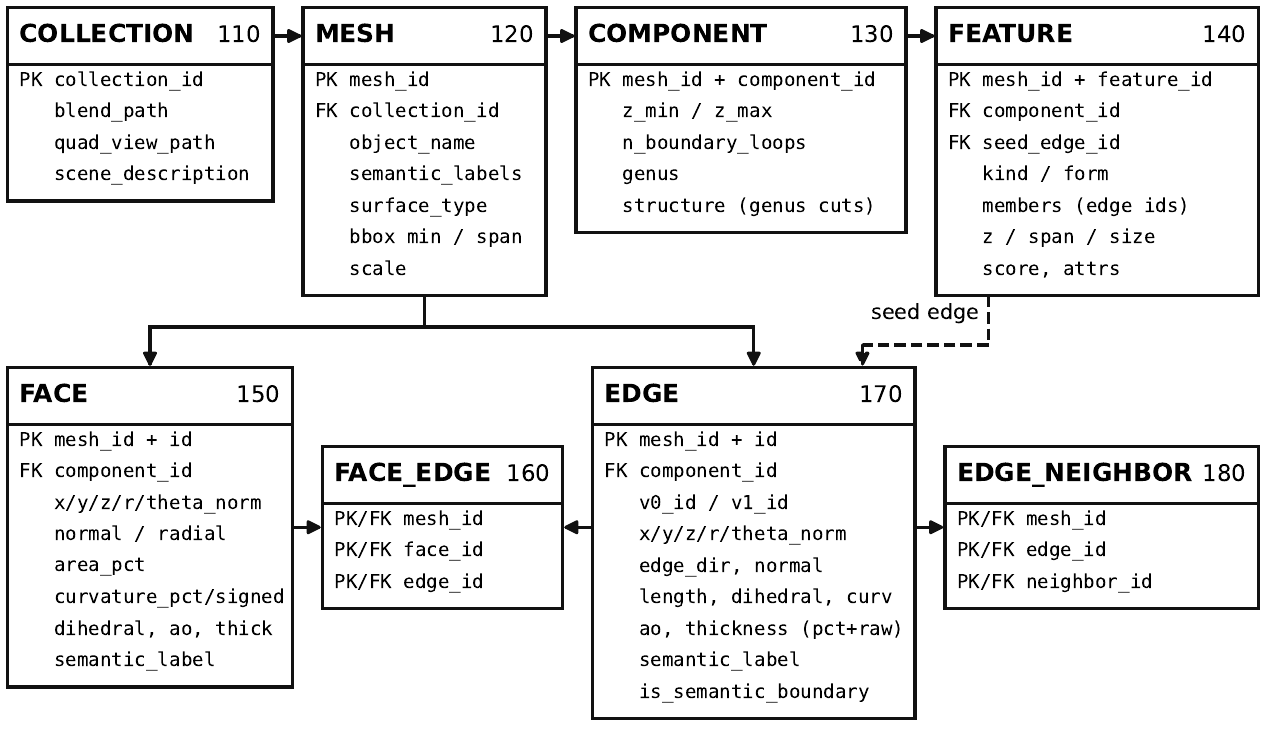}
  \caption{\textbf{Relational mesh representation (\texttt{mesh.db}).}
  PK and FK denote primary and foreign keys. Solid arrows indicate
  one-to-many relations; the dashed arrow links a candidate to its seed edge.
  Geometric signals are stored per face and per edge.}
  \label{fig:relational-schema}
\end{figure}

\mypar{Database model} The eight tables in \texttt{mesh.db}
(Fig.~\ref{fig:relational-schema}) follow the Blender hierarchy:
\texttt{collection} stores asset paths, scene renders, and textual asset descriptions;
\texttt{mesh} stores global information; and
\texttt{component} stores connected-component topology, vertical extent,
and localized genus cuts. The \texttt{face} and \texttt{edge} tables contain
geometric signals and semantic labels, with endpoint, direction, length,
and semantic-boundary fields for edges and area rank for faces.
\texttt{face\_edge} and \texttt{edge\_neighbor} encode incidence and adjacency.
Precomputed candidates reside in \texttt{feature}, with their type, seed
edge, member edges, extent, score, and type-specific attributes. Coordinates use the Blender Z-up frame and isotropic normalization
\((p_a-\min_a)/s\), where
\(s=\max(\operatorname{span}_x,\operatorname{span}_y,
\operatorname{span}_z)\). Suffix \texttt{\_pct} denotes a within-mesh rank in
\([0,1]\), and \texttt{\_raw} retains world-unit or radian values.

\begin{figure}[t]
  \centering
  \includegraphics[width=\textwidth]{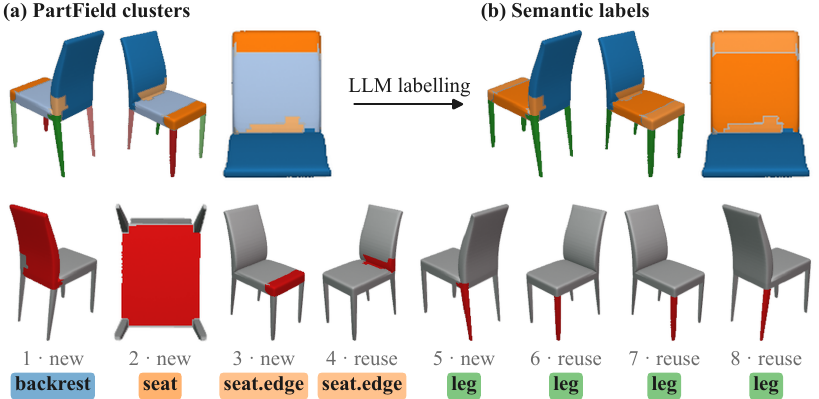}
  \caption{\textbf{VLM semantic labeling}. (a)~\textit{Input}: PartField splits the mesh into $K=8$ clusters based on the mesh semantics and geometry. (b)~\textit{Output}: the VLM produces a final set of labels for each face. The Bottom row illustrates the labeling process.
For each k-th cluster, we render the mesh with the part colored red. Then we prompt the VLM to inspect the image and provide a text label for the selected area. We set the VLM to coin new terms only when strictly necessary, or reuse previously used ones to compose a label vocabulary.}
  \label{fig:vlm-labelling}
\end{figure}

\mypar{Segmentation and semantic labels.}
(Fig.~\ref{fig:vlm-labelling}) A single PartField~\citep{liu2025partfield} inference pass (10 samples per face, batches of 100k) yields an agglomerative face tree, which is cut into $K=8$
clusters by undoing the last $K-1$ merges. Face-level labels are transferred
to the input polygons by nearest normalized centroid, and islets created by the
transfer are merged back so that every cluster is one connected region.
Claude Sonnet~5 first names the object and its surface type, then labels up
to eight clusters in descending face count. Each call receives the original
and highlighted six-view renders and the existing label vocabulary.
Remaining clusters are \texttt{unlabelled}. Faces inherit their cluster label,
edges inherit the label of their left face, and an edge is flagged as a
semantic boundary when its two faces carry different labels.

\mypar{Mesh features.}
Ambient occlusion casts 8 Fibonacci-lattice hemisphere rays of length
$0.3\,d$ ($d$ the bounding-box diagonal) from each edge midpoint, offset by
$0.002$ of the ray length along the averaged normal; the hit fraction is
rank-normalized. Thickness casts 4 inward rays in an $8^\circ$ cone and stores
the median hit distance (NULL for open sheets). Curvature is the signed vertex
angle deficit $2\pi-\sum_i\alpha_i$ (positive on domes, negative on saddles),
averaged over an edge's two endpoints. Dihedral is Blender's signed face angle
for two-manifold edges and zero otherwise. Normals are the unweighted mean of
adjacent face normals; the radial normal is the dot product with the in-plane
radial direction. Rank normalization is $\mathrm{rank}/(N-1)$ with min-rank
ties; edge length is normalized by the maximum length. All ray queries share
one world-space BVH.

\mypar{Precomputed geometric features.}
\emph{Ring candidates} are connected runs of transition edges (semantic
boundary, or curvature or dihedral rank $\geq 0.9$) of at least 6 edges. Each
run is fitted with its own PCA plane and accepted as a ring when
thickness/radius $\leq 0.15$ and in-plane angular coverage $\geq 0.9$; the
most tangent edge is stored as the seed. \emph{Curvature ridges} are connected
runs of dihedral rank $\geq 0.92$ with at least 6 edges. \emph{Genus necks}
come from the per-component Euler characteristic and a localized handle or
tunnel cut. \emph{Symmetry planes} test the three axis-aligned planes through
the bounding-box center: up to 1500 edge midpoints are reflected and matched
to their nearest neighbor in a grid of cell size $0.02\,d$; a plane is accepted
when the median reflection error is below $0.06\,d$ and at least 40\% of points
match within two cells, and the plane with the lowest error is stored.

\subsection{Baselines}
\label{app:baselines}

Blender Smart
UV uses angle limit $66^\circ$ and island margin $0.03$. \textsc{xatlas} uses
default chart options. OptCuts runs headless with $\lambda=0.999$ and
distortion bound $4.1$, seed 0. \textsc{PartUV} uses all default values at seed $42$, distortion
threshold $1.25$, and the same PartField sampling as our preprocessing.
FlattenAnything uses 10{,}000 points and 10{,}000 iterations, reseeded per
object. Every baseline is given a 40-minute budget; samples that
time out are recorded as failed.

\subsection{Parameterization Algorithm}
\label{app:parameterization}
Parameterization algorithm's quality is mesh-dependent: for instance conformal performs best on hard-surface objects, while ABF++ performs better on organic ones.
We choose the parameterization algorithm based on the distribution of dihedral-angle edges: the framework selects LSCM (hard-surface specific) if the percentage of such edges is above a given threshold; otherwise, Min. Stretch is used.  In addition, the framework may fallback from Min. Stretch to ABF++ if the current seams do not reach the user-specified distortion thresholds.

\subsection{Metrics}
\label{app:evaluation-metrics}

\mypar{Chart count.}
Two faces are in the same chart when they share a full UV edge; the number of
connected components in the UV atlas is the chart count. 

\mypar{Seam Length.} UV seam length is the total UV
length of chart boundary edges. 3D seam length is the 3D length of the cut
edges, each physical cut counted once, divided by the total 3D area. The
non-disk fraction
is the share of charts whose Euler characteristic differs from one (annuli,
handles, closed components).

\mypar{Chart shape.} Following \cite{ma2026meshtailor}, for a chart with UV area $a$, perimeter $p$, and convex-hull area $h$,
compactness is $4\pi a/p^2$ and convexity is $\min(1,a/h)$; both are averaged
over charts, skipping charts with zero area or perimeter. For jaggedness, each
closed UV boundary loop is resampled at spacing $0.005$ (8 to 4096 points) and
we average the second difference $\|q_{i-1}-2q_i+q_{i+1}\|$ of consecutive
samples $q_i$.

\mypar{Seam visibility.}
Occlusion at a surface point is measured by casting $8$ cosine-weighted
hemisphere rays and taking the fraction that hit the mesh again: $0$ on an open
surface, $1$ deep inside a crevice. We evaluate it at $N=1000$ points placed
along the seams, spaced evenly by arc length so that each sample stands for the
same amount of seam, giving values $s_1,\dots,s_N$. The same measurement on
$N$ points spread over \emph{all} edges of that mesh gives $e_1,\dots,e_N$,
which describe how occluded that surface is in general.

Reporting the mean of the $s_i$ would mostly report the shape: a convex object
occludes nothing, so the mean is near zero wherever the seams run. We therefore
score the seams against their own mesh, by comparing every seam value with every
reference value and taking the fraction of comparisons in which the seam point
is the more occluded one, ties counted as half:
\[
  \mathrm{AO}=\frac{1}{N^2}\sum_{i=1}^{N}\sum_{j=1}^{N}
  \Big[\mathbf{1}\{s_i>e_j\}+\tfrac12\,\mathbf{1}\{s_i=e_j\}\Big].
\]
Being a fraction of comparisons, AO is in $[0,1]$ by construction and needs no
further normalization: $1$ means every seam point is more hidden than every
point of the surface, $0.5$ that seams are indistinguishable from the surface at
large, and below $0.5$ that they run along exposed regions. It is therefore
comparable across meshes, and stays defined on a convex shape, where all
$2N$ samples are zero, and AO is exactly $0.5$.

\mypar{Area distortion.} We follow the same  implementation as~\citep{wang2025partuv}. A face $f$
is kept if $\mathrm{area}_{3D}(f)$ is at least $10^{-4}$ of the mean 3D face
area and $\mathrm{area}_{2D}(f)>0$. Its area ratio is normalized by the
atlas-wide ratio, then folded and capped at 10:

\[
s_f=
\frac{\mathrm{area}_{2D}(f)\sum_g \mathrm{area}_{3D}(g)}
     {\mathrm{area}_{3D}(f)\sum_g \mathrm{area}_{2D}(g)},
\qquad
\mathrm{area\_dist}(f)=
\min\!\left(10,\max\!\left(s_f,\frac{1}{s_f}\right)\right).
\]

Thus $d(f)\ge1$, equal to one for an isometric face, and stretching or
shrinking by the same factor scores the same. Overall area distortion is the
plain mean of $d(f)$ over kept faces; the cap at 10 follows the original code
and keeps a few degenerate faces from dominating the mean. Worst-chart area
distortion recomputes the same mean per chart, with the sums and kept faces
restricted to that chart, and takes the largest.

\mypar{Angular distortion.}  We follow the same  implementation as~\citep{wang2025partuv}. 
Per face, let $e_u$ and $e_v$ be the 3D directions along which $u$ and $v$
increase, and $\theta$ the angle between them. Then
\[
cos\theta(f)=\frac{e_u\cdot e_v}{|e_u|,|e_v|},\qquad Q_{\mathrm{ang}}=1-\frac1F\sum_f\big|\cos\theta(f)\big|
\]
over faces with a non-degenerate UV triangle. It is one when every face is
mapped with perpendicular axes; it tests orthogonality only, not equal
stretch along the two axes.

\mypar{Conformal distortion.}
Per face, take the $2\times2$ Jacobian from the tangent plane to UV with
singular values $\sigma_1\ge\sigma_2$. Conformal distortion is the mean of
\[
 K=\sigma_1/\sigma_2
\]
over
\[
 \mathcal V_K=\{f\in\mathcal V_A:
 \det[u_{f,1}-u_{f,0}\;\;u_{f,2}-u_{f,0}]>10^{-12}\},
\]
the nondegenerate, orientation-preserving faces. Thus flipped and collapsed UV triangles are omitted while every
atlas still contributes one aggregate. One is ideal, and lower is better.

\mypar{Texel spread.}
With $t(f)=\log_2\big(\mathrm{area}_{2D}(f)/\mathrm{area}_{3D}(f)\big)$ the
per-face log texel density, texel spread is the standard deviation of $t(f)$
weighted by $\mathrm{area}_{3D}(f)$, in octaves. It is zero when every face
gets the same texture resolution and does not depend on the global UV scale.

\subsection{Measuring Token count}
The observation tokens in Fig.~\ref{fig:resolution-scaling}(right) are the total text returned by tool calls during one run, while \emph{OBJ serialization} is the byte size of the mesh's OBJ files.  Both are converted to tokens at four characters per token.  Overall, for 100 meshes, the median run reads
36k observation tokens (min 10k, max 175k), against 230k tokens for the
serialized mesh, and the two scale differently with resolution: the
serialized size grows linearly with polygon count (Spearman $\rho=+0.99$,
fitted slope 1.06 per decade in log--log), whereas observation tokens are
nearly flat ($\rho=+0.15$, slope 0.05 per decade). The largest per-turn context, including cached prompt and all prior tool results, has a median of 157k tokens and a maximum of 503k.

\section{Extended Results}
\begin{figure}[t]
    \centering
    \includegraphics[width=1\linewidth]{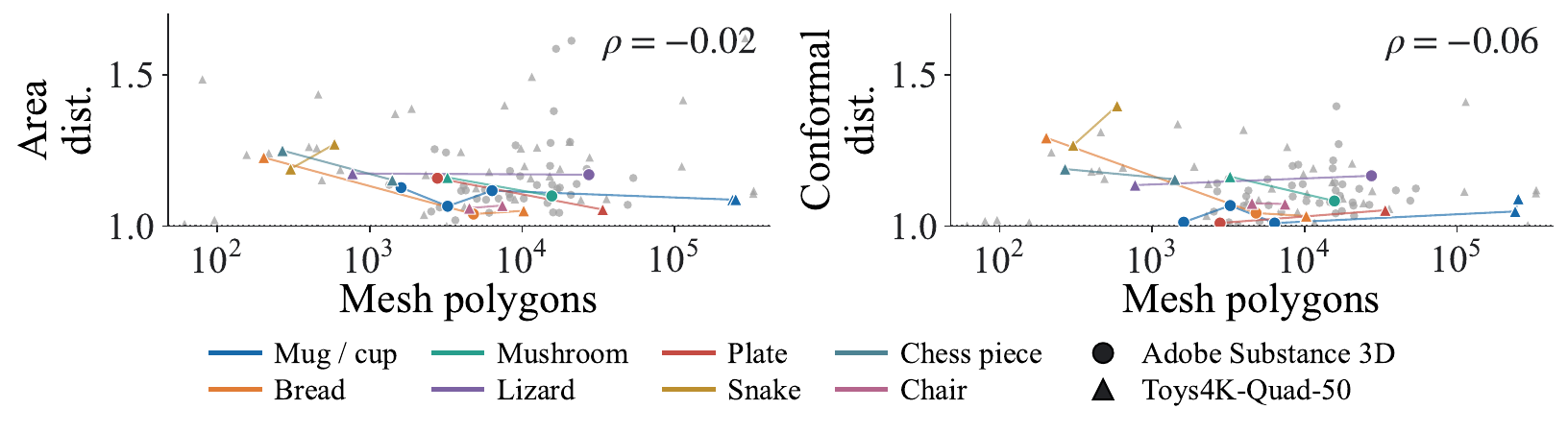}
    \caption{\textbf{Distortion performance scaling over mesh size.} \method distortion performance does not degrade over larger meshes.}
    \label{fig:distortion_plots}
\end{figure}

We extend Fig.~\ref{fig:resolution-scaling} of the main paper with pointwise distortion plots in Fig.~\ref{fig:distortion_plots}. Our method scales well to larger meshes, and distortion performance is uncorrelated with mesh polygon size.

\begin{figure}[t]
  \includegraphics[width=\linewidth]{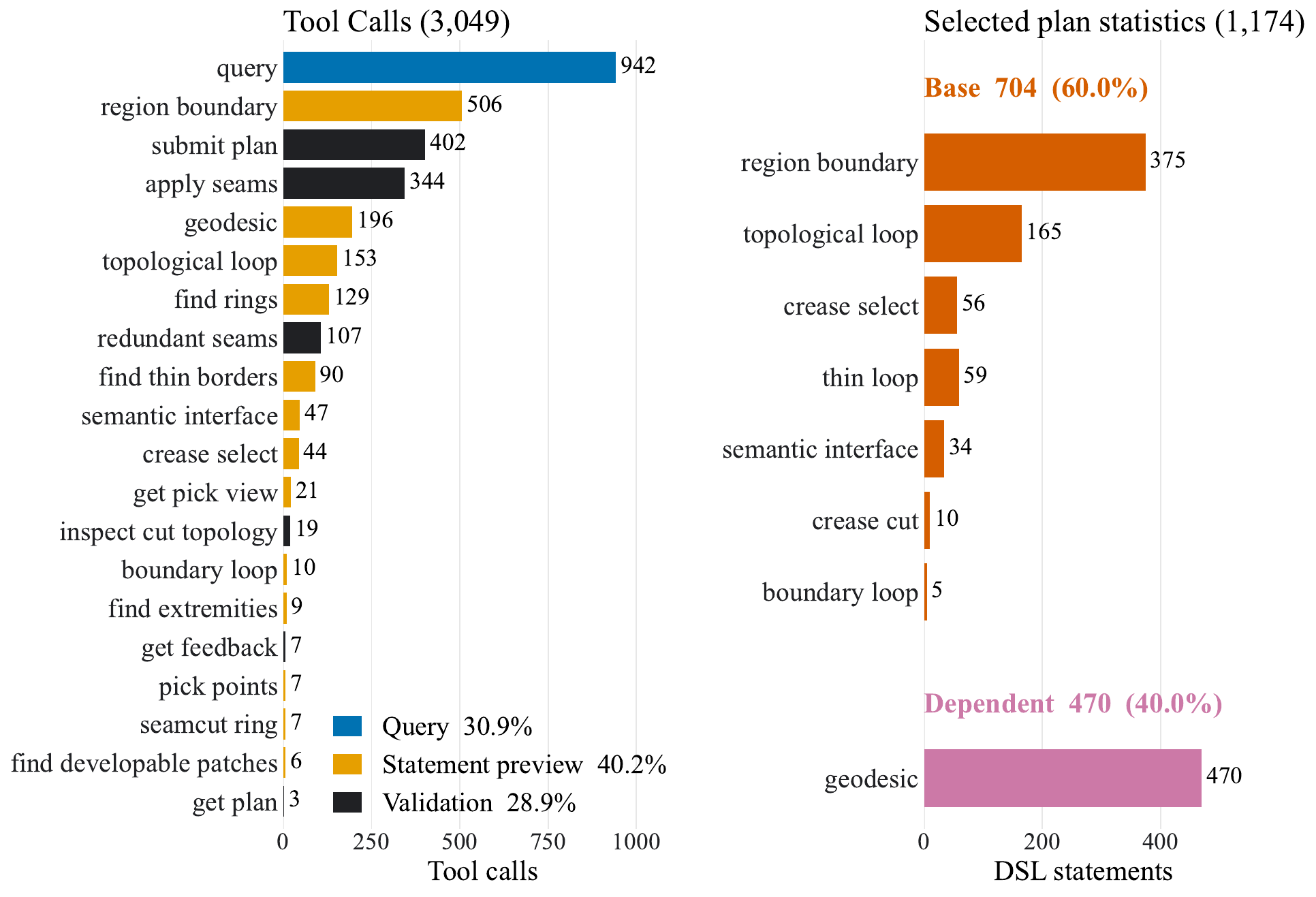}
\caption{\textbf{Tool use across both evaluation sets (100 runs).} \emph{Left:} Distribution of the 3{,}049 tool calls by tool and call kind. \emph{Right:} Distribution of the 1{,}174 DSL statements in the 100 selected programs, grouped by the type of edge selection returned.}
  \label{fig:tool-analysis}
\end{figure}

\mypar{Tool usage statistics.} Fig.~\ref{fig:tool-analysis} counts the tool calls of the selected plans across the evaluation benchmark. We observe over $3,049$ tool calls, of which 30.9\% are queries, 40.2\% are statement previews, and 28.9\% are validations. In addition, we also report 55.8\% base DSL statements and 44.2\% dependent types.

\begin{figure}[t]
    \centering
    \includegraphics[width=0.9\linewidth]{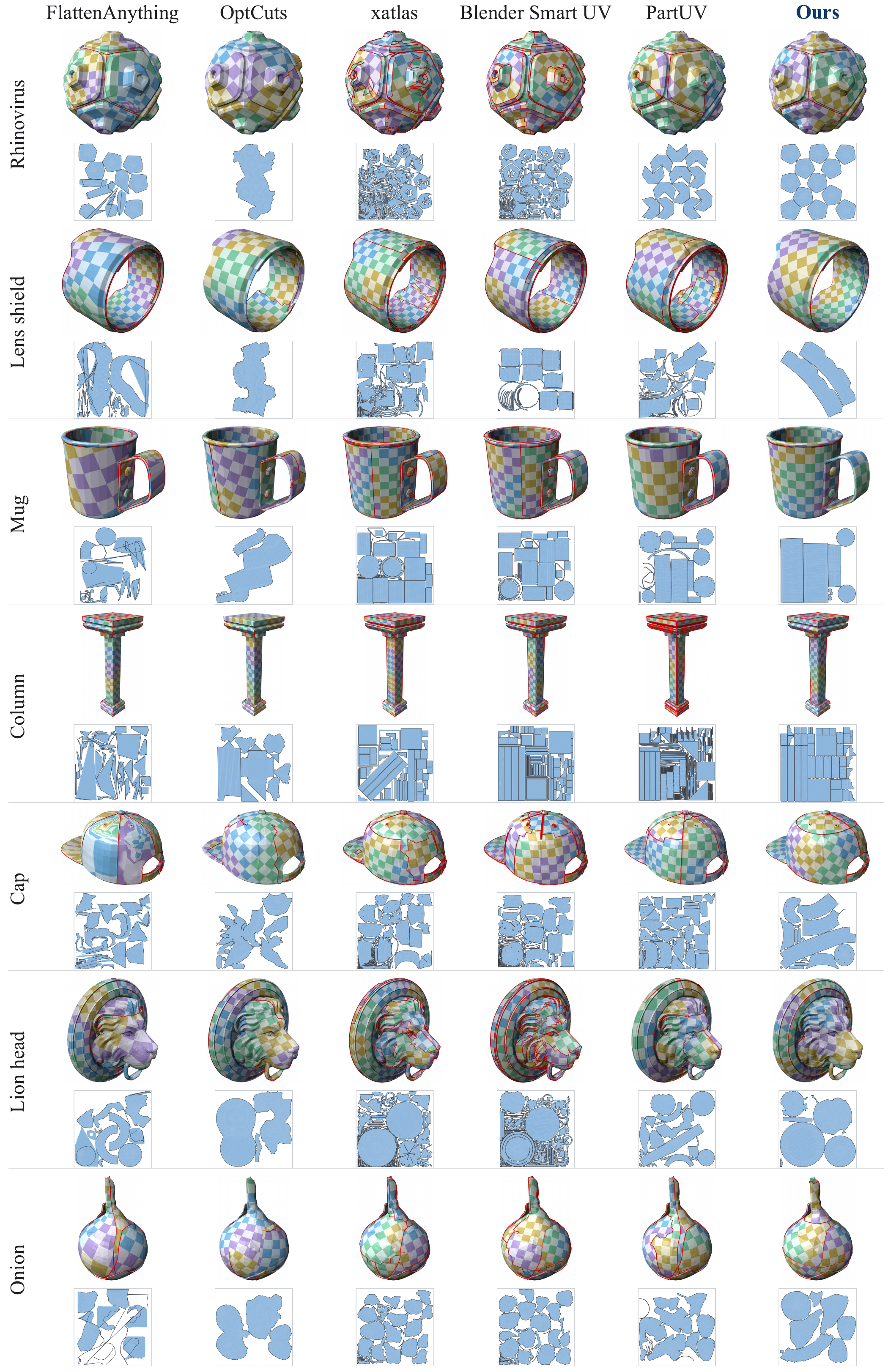}
 \caption{\textbf{Qualitative comparison on \textsc{Substance-50}.} For each
  asset, the top row shows a UV checkerboard and generated seams in red, while the bottom row shows the parametrized UV atlas.}
    \label{fig:qualitative_substance}
\end{figure}

\mypar{Qualitative results} of main paper Fig.~\ref{fig:qualitative} are extended  in Fig.~\ref{fig:qualitative_substance}.

\end{document}